\documentclass{article}

\usepackage{arxiv}
\usepackage[T1]{fontenc}
\usepackage{lineno}
\usepackage{fix-cm}
\usepackage{graphicx}
\usepackage{multirow}
\usepackage{amsmath,amssymb,amsfonts}
\usepackage{amsthm}
\usepackage{array}
\usepackage[title]{appendix}
\usepackage{natbib}
\usepackage{xcolor}
\usepackage{booktabs}
\usepackage{listings}
\usepackage{hyperref}%\autoref{}
\usepackage{arydshln}
\usepackage{fancyvrb}
\usepackage[mathscr]{eucal}
\usepackage[figuresright]{rotating}
\usepackage{textcomp}%
\usepackage{manyfoot}%
\usepackage{algorithm}%
\usepackage{algorithmicx}%
\usepackage{algpseudocode}%
\usepackage{setspace} 
\usepackage{indentfirst}
\newcommand{\re}{\textcolor{black}}

\title{DeepGreen: Effective LLM-Driven Greenwashing Monitoring System Designed for Empirical Testing \textemdash Evidence from China}

\author{
  \textbf{Congluo Xu}\thanks{First Author. Email \texttt{xucongluo@stu.scu.edu.cn}}\\
  Business School\\ Sichuan University\\
  Chengdu, 610065
  \\[2ex]
  \textbf{Ziyang Li}\thanks{Corresponding Author. Email \texttt{lzy\_feng@scu.edu.cn}}\\
  Business School\\ Sichuan University\\
  Chengdu, 610065
  \And
  \textbf{Jiuyue Liu} \\
  School of Economics\\ Sichuan University\\
  Chengdu, 610065
  \\[2ex]
  \textbf{Chengmengjia Lin}\\
  Business School\\ Sichuan University\\
  Chengdu, 610065
}

\begin{document}

\maketitle

\begin{abstract}
Motivated by the emerging adoption of Large Language Models (LLMs) in economics and management research, this paper investigates whether LLMs can reliably identify corporate greenwashing narratives and, more importantly, whether and how the greenwashing signals extracted from textual disclosures can be used to empirically identify causal effects. To this end, this paper proposes DeepGreen, a dual-stage LLM-Driven system for detecting potential corporate greenwashing in annual reports. Applied to 9369 A-share annual reports published between 2021 and 2023, DeepGreen attains high reliability in random-sample validation at both stages. Ablation experiment shows that Retrieval-Augmented Generation (RAG) reduces hallucinations, as compared to simply lengthening the input window. Empirical tests indicate that ``greenwashing'' captured by DeepGreen can effectively reveal a positive relationship between greenwashing and environmental penalties, and IV, PSM, Placebo test, which enhance the robustness and causal effects of the empirical evidence. Further study suggests that the presence and number of green investors can weaken the positive correlation between greenwashing and penalties. Heterogeneity analysis shows that the positive relationship between ``greenwashing - penalty'' is less significant in large-sized corporations and corporations that have accumulated green assets, indicating that these green assets may be exploited as a credibility shield for greenwashing. Our findings demonstrate that LLMs can standardize ESG oversight by early warning and direct regulators’ scarce attention toward the subsets of corporations where monitoring is more warranted.
\end{abstract}

% keywords can be removed

\section{Introduction}\label{secIntroduction}
\re{Against a backdrop of sharpening international commitment to sustainable development and environmental protection, the practice of corporate ``greenwashing''` has risen to prominence as a fiercely contested issue \citep{walker2012harm, szabo2021perceived}.} \re{``Greenwashing'' typically denotes the strategic exaggeration or misrepresentation of}  their environmental protection efforts in promotional materials, while their actual approach did not involve or meet the standards of sustainable development \citep{lyon2011greenwash, de2020concepts, ziolo2024literature}. \re{Under this definition, in contrast to verifiably false disclosures — which regulators can refute by comparing statements against observable facts — corporations can invoke open-ended slogans such as ``carbon peak'' or ''green development'' without committing to measurable actions \citep{lyon2011greenwash, liu2023greenwashing}. Precisely because such vague assertions are disconnected from verifiable benchmarks, they circumvent conventional enforcement mechanisms and pose a primary detection problem for external stakeholders \citep{kim2015greenwash}.}

\re{The elusive nature of greenwashing mentioned above stems from its utilization of human psychology and information processing mechanisms.} By strategically emphasizing environmentally related terminology, companies construct a superficial image of environmental responsibility \citep{cai2016corporate, jones2019rethinking}. \re{This strategy has created a huge obstacle, making it more difficult for analysts and consumers to establish a rational understanding of the company's green image from its written materials, and leaving behind a sensory understanding of the company's ``environmental commitment'' \citep{parguel2015can, seele2017greenwashing, acheampong2024social}.} Such strategic manipulation not only undermines the credibility of corporate sustainability claims but also complicates accountability efforts regarding environmental impact.

\re{The reason for studying the greenwashing behaviour of corporations is largely to prevent potential environmental and social harm caused by their actions \citep{lyon2015means,li2023effects}. However, how to transform the textual definition of greenwashing from literature into quantifiable criteria, and furthermore, whether the greenwashing information detected from corporate financial reports can serve as a proxy for detecting corporate environmental penalties, are pre-questions that worth answering.}

In the face of these \re{questions, previous studies have developed two mature measurement methods. \cite{cao2022carbon, xu2023unveiling} draw inspiration from \cite{ashforth1990double}, construct the ratio of verbal (symbolic) and actual (substantive) disclosure quantities to measure greenwashing. \cite{delmas2011drivers, kim2015greenwash, yu2020greenwashing} subtract the standardized value of ``ESG performance score'' from the standardized value of ``ESG disclosure score'', and consider those values greater than zero as greenwashing.} \re{Because both constructs hinge on measuring corporations' ESG disclosure—whose evaluation of information is inherently more ambiguous and harder to delimit than ESG performance—reliance on third-party ratings inevitably places researchers in a passive position \citep{christensen2022corporate}. Meanwhile, with the significant increase in the length of text disclosure in corporate annual reports in recent years \citep{dyer2017evolution, clapham2023policy}, the most primitive and trustworthy method of} manual review often fall short due to the sheer volume of data and the sophistication of greenwashing tactics \citep{bernini2024measuring, martin2016managers}. \re{In order to efficiently measure the disclosure level of corporations while reducing institutional bias, text analysis methods have begun to be widely applied \citep{bochkay2023textual}.}

\re{Researchers initially choose statistics on the frequency of green related words as a proxy for corporate ESG disclosure, but these methods are inevitably subject to serious systemic bias caused by corporate greenwashing \citep{bingler2022cheap}. Supervised deep learning classifiers (e.g. RNN \citep{zhou2024threat}) require a large number of manually annotated samples to ``learn'' the ability to recognize greenness before processing the task of identifying greenness keywords. This kind of manual intervention greatly limits the large-scale application of such deep learning models. The two major disadvantages of the above text analysis methods are the high level of manual intervention and the relatively basic level of natural language processing. Therefore, it is difficult to demonstrate whether these ``text fragments'' can be transformed into qualified quantifiable greenwashing indicators. }

Recent advancements in artificial intelligence, especially Large Language Models (LLMs), offer promising solutions to these limitations \citep{boedijanto2024potentials}. \re{LLMs, with their ability to simulate manual text ``reading'' and process a large amount of context in a very short period of time,} provide a powerful means of analysing financial statements and corporate disclosures. \re{By utilizing this ability,} LLMs can not only identify specific keywords related to environmental claims but also contextualize these terms within the broader narrative of a company's sustainability efforts \citep{zou2025esgreveal}. \re{Nevertheless, there is relatively few studies on whether the greenwashing judged by LLM based on the original definition of "Oral - Implemented" can become a qualified proxy as corporate greenwashing, both technically and empirically.}

\re{To investigate whether LLMs really satisfy the two significant pre-questions mentioned above on greenwashing studies,} this paper introduces DeepGreen, an innovative dual-layer framework that \re{simulating human judgment for corporate greenwashing.} The \re{marginal} contributions of our paper are listed as follows:

(1) \re{This paper constructs a corporate greenwashing detection framework that integrates LLMs with a simulated manual-review protocol to extract green disclosure items from annual reports.}

(2) \re{Technically, through random sampling verification of the judgment results in the two stages of ``exploring potential green keywords'' and ``verifying whether keywords are implemented'', it has been confirmed that DeepGreen framework can satisfy the requirements of the judgment tasks in both stages.}

(3) \re{Empirically, we explore whether green disclosure results on annual reports obtained through Deepgreen can serve as a valid proxy for corporate greenwashing. The empirical results show that this proxy is significantly positively associated with subsequent environmental penalty. This relationship survives IV, PSM, and placebo tests.}

(4) \re{This paper provides affirmative evidence on the reliability of an unsupervised LLM-Driven framework for detecting corporate greenwashing, and the heterogeneity analyses highlight corporate categories that regulators and agencies may prioritize.}

The reminder of this paper are arranged in the following order. \re{We first systematically review the mainstream criteria for defining corporate greenwashing and appraise the construction methods and proxy validity of conventional greenwashing indicators (\S~\ref{secRelatedwork} Related Work).} \re{Next}, we represent DeepGreen's framework overview and its technical details (\S~\ref{secFramework} Method Explanation). \re{We then describe the experimental setting—covering the dataset, task templates, model selection, empirical results, validation procedures, and ablation analyses (\S~\ref{secExperiment} Experiment Details). Thereafter, we examine the association between corporate greenwashing and environmental penalties (\S~\ref{secEmpirical} Empirical Test). Finally, we integrate the foregoing analyses to address two foundational questions (\S~\ref{secConclusion} Discussion and Conclusion): whether LLMs can reliably detect greenwashing in annual reports and whether textual greenwashing signals serve as valid proxies for corporations’ greenwashing conduct.}

\section{Related Work}\label{secRelatedwork}
\re{Most current study on greenwashing behaviour assumes that their identification strategies for corporate greenwashing behaviour are effective, as these identification strategies are naturally derived from the theoretical concepts of \cite{delmas2011drivers, lyon2011greenwash, seele2017greenwashing, de2020concepts}. However, whether LLMs backing complex emergence mechanism can still apply the same quantitative guidelines as before and reliably complete tasks based on this set of criteria is still controversial and unclear in related studies.} 

\re{To conduct a focused investigation, this section critically synthesizes the extant related work and try to clarify three main questions. Firstly, we will organize the conceptual definition of the corporate greenwashing and answer why we choose annual report data as the analysis object.} (\S~\ref{2.1}). \re{Subsequently, the mainstream measurement criteria for corporate greenwashing will be reviewed, and the criteria used in this paper and its corresponding reasons will be provided afterwards} (\S~\ref{2.2}). \re{Finally, this paper attempts to address the core technical issues that have been overlooked in previous related studies, providing explanations and support for our proposed technological path.} (\S~\ref{2.3}).

\subsection{Theoretical Definition and Ideal Carrier for Corporate Greenwashing}\label{2.1}
\re{Greenwashing behaviour originates from the market's progressively attention to the greening and sustainability profile of corporations. It intends to manipulate the information asymmetry between the market and the corporation, in order to move towards a favorable situation for the company. \cite{yue2023media} argues that as capital-market scrutiny of corporate green reputations intensifies, corporations face heightened incentives to exploit information asymmetries by strategically overstating their environmental performance, thereby securing cheaper capital and reputational rents.}

\re{This viewpoint has evolved continuously from early research. \cite{delmas2011drivers} describes corporate greenwashing as misleading consumers' perception of the company's environmental performance or the environmental benefits. \cite{lyon2011greenwash} classifies greenwashing as the selective disclosure of information about a company's environmental or social performance to create an overly positive but false image. \cite{seele2017greenwashing} further elaborates that greenwashing is a symbolic strategy adopted by corporations to obtain or maintain their environmental legitimacy. \cite{de2020concepts} systematically compiles relevant studies from the past decade, condensing greenwashing into any dissemination behaviour that misleads people into forming overly positive beliefs about organizational environmental performance, practices, or products. From this, it can be seen that corporate greenwashing is a publicize behaviour that must rely on certain carriers \citep{lyon2015means, marquis2016scrutiny, gorovaia2025identifying}.}

Today's \re{corporate annual} reports have undergone a transformation from simple financial reports \re{into instruments for promotion and presentation to investors \citep{mohanram2025does}. Unlike targeted materials such as CSR reports and ESG responsibility statements, which may only be of interest to ESG experts or environmentalists, all market investors need to review the company's annual report. In addition, CSR and ESG reports are voluntarily issued and non-mandatory audited, however, they are usually read by professionals in related fields \citep{cohen2015nonfinancial}, thus greenwashing is easier to be detected and the cost of greenwashing is much higher. While annual reports must go through the audit process, so its is naturally considered a ``more reliable'' disclosure material \citep{li2025annual}. What's more, the focus of audit is actually on the compliance of accounting records and financial violations instead of ESG \citep{grigoras2024importance}, which gives the annual report a lot of room for greenwashing. Therefore, corporate annual reports become the ideal greenwashing carrier with the widest audience and built-in ``reliability'' gain.} 

\re{Following the theoretical definition of greenwashing in previous research, we deem that corporate greenwashing should be measured through the presentation effect of its carrier. Moreover, as China issued the new annual report format in 2021, which includes ESG as a separate chapter, we reasonably choose annual report data as our analysis object.}

\subsection{Measurement Criteria and Indicator for Corporate Greenwashing}\label{2.2}
\re{The vast majority of recent studies on corporate greenwashing behaviour measures the consistency of a company's words and actions \citep{wei2023does, mu2023greenwashing, wang2025construction}. Grounding the analysis in institutional theory and signal transmission theory, \cite{lubloy2025quantifying} demonstrates that, because legitimacy rather than efficiency governs organisational survival, and because external investors observe only noisy signals while firms retain perfect knowledge of their true environmental capabilities, decoupling outward-facing green symbols from unchanged core operations is the defining mechanism through which greenwashing emerges.
}

\re{When constructing a corporate word and action consistency index as a proxy for corporate greenwashing behaviour, the two most mainstream approaches are vastly different — one relies on the corporation's textual disclosure, while the other relies on external market ratings.} 

\re{\cite{cao2022carbon, xu2023unveiling, gorovaia2025identifying} resort to text analysis methods to distinguish corporate promotional green keywords $X$ and actual performance keywords $Y$, and construct the relative indicator: 
\begin{equation}\label{gw1}
    GW=\dfrac{\sum X}{\sum Y+1}
\end{equation}
However, one question about this method is whether the green performance disclosed in the company's text is the true green performance \citep{calamai2025corporate, sneideriene2025uncovering}. This has raised concerns about insufficient practical basis.}

\re{Another measurement path adopted by \cite{yu2020greenwashing, li2024does, tan2024faking} is the difference between the ESG disclosure score issued by the institution and the ESG performance score:
\begin{equation}
    GW=\left(\dfrac{ESG_{Dis}-\overline{ESG_{Dis}}}{\sigma_{Dis}}\right)-\left(\dfrac{ESG_{Per}-\overline{ESG_{Per}}}{\sigma_{Per}}\right)
\end{equation}
This measurement method is widely used due to the fact that the data used are all institutional ratings, resulting in lower acquisition costs. However, it is worth noting that although institutions are relatively authoritative and reliable in evaluating actual performance, their criteria for measuring and rating disclosure levels often may not be completely consistent with the focus of researchers, leading to the possibility of systematic bias \citep{berg2022aggregate}.}

\re{Although some studies have linked text analysis indicators to actual performance \citep{mateo2022international, shi2024effect}, most of them are related to specific indicators such as carbon emissions and sewage discharge, and the comprehensive environmental performance of enterprises has not been effectively evaluated, which may also lead to bias. For this purpose, we use the text analysis results of corporate annual reports to obtain the green disclosure level of corporate texts for research purposes, and also select the environmental rating from authoritative institutions' ESG ratings as a comprehensive and representative level of corporate green performance. Meanwhile, due to the significant subjectivity in scoring text results \citep{bridgeman2012comparison, amorim2018automated}, we employ the word frequency ratio itself rather than the scoring results.}

\subsection{Technical Review of Text Analysis for Green Disclosure}\label{2.3}
The organization and utilization of textual information require huge integration costs, \re{which means that manual review is never a best appropriate solution. As a consequence, the research on corporate green disclosure begins with the frequency statistics of green keywords. Researchers pre-set a green keyword library and identify words or language materials that meet the criteria from the text, which is achieved through Bag-of-Word (BoW) \citep{harris1954distributional} and Lexicon-based Methods \citep{taboada2011lexicon}. But due to ignoring the word order and grammar of the text \citep{kroon2024advancing}, word frequency cannot distinguish between valid and invalid concepts in disclosure, and was quickly replaced by ratio (\autoref{gw1}) on corporate greenwashing.}

\re{However, distinguishing the valid part in green keywords to identify potential corporate greenwashing is also time-consuming task, and most existing studies train their supervised deep learning classifier models to distinguish \citep{zhou2024threat, cheng2025impact, wang2025construction}. This type of technical approach not only requires high requirements for pre-labeled data, but may also suffer from overfitting and insufficient generalization performance \citep{rohlfs2025generalization}. Meanwhile, it is not a natural transformation of the recognition strategy under the definition of greenwashing behaviour in theoretical studies. The direct reason is that these deep learning models above are only classifiers rather than a natural-language-like interactive system, and naturally cannot transfer human recognition strategies and ideas into greenwashing dentifying tasks \citep{ong2025towards, persakis2025greenwashing}.}

With the enhancement of practical computility, some theoretical models of natural language processing (NLP) have been implemented. Large language models (LLMs) based on Transformer architecture \citep{vaswani2017attention} \re{demonstrate the potential of naturally processing textual data according to the logic of human language.} Some of the representative ones are GPT-4 \citep{achiam2023gpt}, LLaMA-3 \citep{dubey2024llama}, \re{Deepseek-V3 \citep{liu2024deepseek}, KIMI-K2 \citep{team2025kimi}. LLMs show the potential of perfectly transferring manual recognition and calibration to automated programs.} However, when we try to utilize LLMs for measuring the effectiveness of corporate green disclosure, there're now facing several obstacles.

Firstly, previous studies indicates that LLMs cause hallucinations, which might reduce task performance \citep{zhang2025siren, hua2024limitations}. Secondly, the contradiction between qualitative unstructured ESG information extraction and the demand from quantitative and empirical study creates tremendous challenges in adapting to a uniform framework for studying greenwashing \re{behaviour}. \re{Thirdly,} semantic understanding and generation capabilities of LLM can be utilized as indicators related to greenwashing \citep{bronzini2024glitter}, but the lack of quantitative criteria for defining greenwashing  makes ``greenwashing detective'' difficult to establish across the various \re{corporations.} 

Due to the aforementioned difficulties, this paper proposes ``DeepGreen''. \re{This framework provides three major functions for our research. Firstly, we will attempt to fully transfer the enterprise greenwashing identification criteria proposed by \cite{seele2017greenwashing} to an LLM-Driven framework to explore whether LLMs can effectively identify the true level of green disclosure of corporations. Secondly, we will validate the effectiveness of enhancing contextual texture and employing RAG in alleviating LLM hallucinations and improving recognition through ablation experiments. Thirdly, we refer to \cite{xu2023unveiling} to construct an indicator of corporate green disclosure, which serves as one of the evaluation variables for identifying greenwashing. Based on the green performance variable used by \cite{yu2020greenwashing}, we will construct a binary indicator of corporate greenwashing for subsequent empirical analysis.}

\section{Framework}\label{secFramework}
\subsection{Overview}

\begin{figure}[!htbp]
    \centering
    \caption{\centering Overview of DeepGreen Framework.}
    \includegraphics[width=1\linewidth]{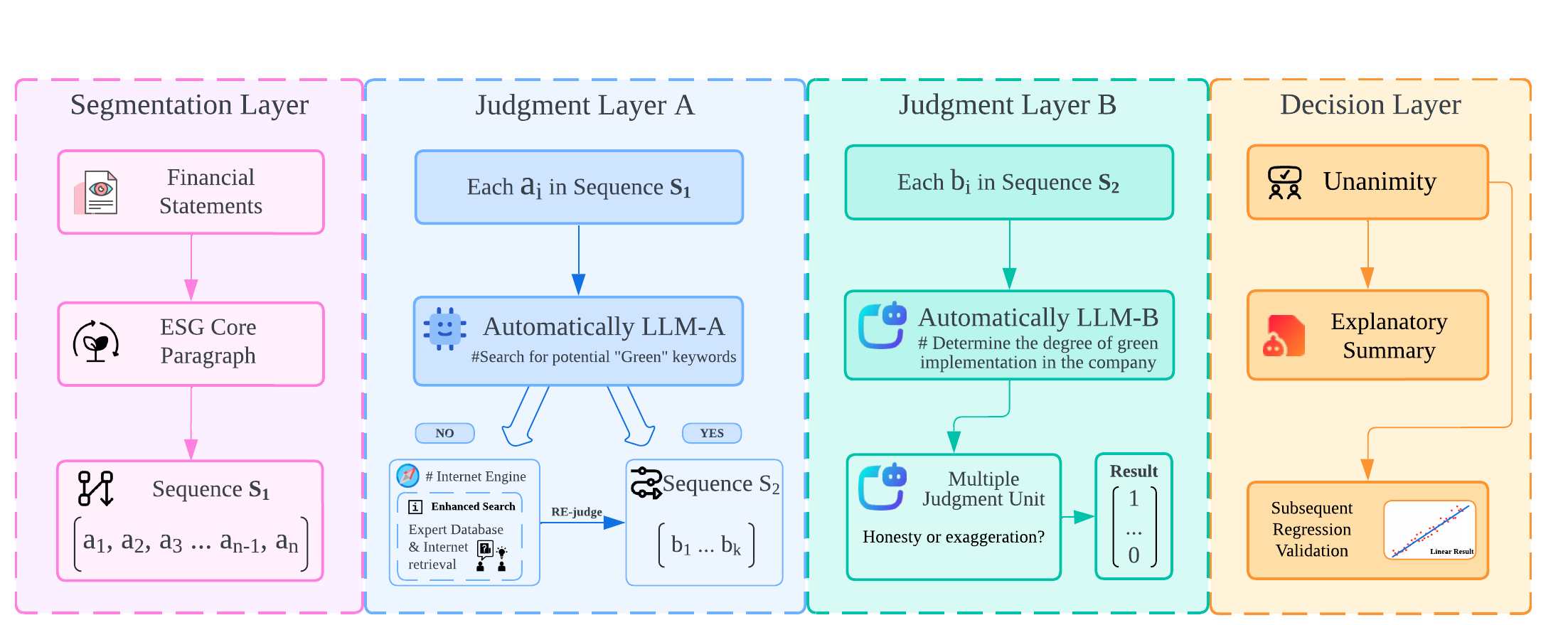}
    \label{fig1}
\end{figure}

\re{We schematically summarize the DeepGreen workflow in \autoref{fig1}, which presents a four layer framework engineered to construct a valid green disclosure level that filters out the noise inherent in symbolic green disclosure.} \re{In the Segmentation Layer, we employ Jieba and HIT Stopword List, spliting ``Environmental Information'' section from annual report into words and construct a unique word sequence $\mathbf{S_1}$.} Judgment Layer A employs LLM to extract ``Green'' keywords in each element of $\mathbf{S_1}$, delivering these elements to sequence $\mathbf{S_2}$. In Judgment Layer B, \re{LLM determines whether the element in b is a symbolic disclosure or a substantial improvement disclosure based on the} elements of $\mathbf{S_2}$. \re{Decision Layer summarizes the judgment results of the previous layer according to the classification of each corporation, and finally provide the effectiveness indicators of corporate green disclosure.} Specifically, the critical layers for measuring the effectiveness of green disclosure are Judgment Layer A and Judgment Layer B, where we utilize a hybrid approach of ``\re{\textbf{Expert Pre-set}} $\rightarrow$ \textbf{Enhanced Retrieval} $\rightarrow$ \re{\textbf{Judgment}}'' to enhance the performance. \re{In contrast to the traditional recognition pipelines, this design significantly reduces manual supervision, and the detailed comparison will be provided in \autoref{comparison}.}

\subsection{Green Disclosure Keywords}
\re{To measure corporations' substantive green disclosure, we must ensure that our method covers the parts that have been achievable in previous studies — extracting green keywords from $\re{\mathbf{S_1}}$} as comprehensively as possible. \re{Traditional practice is to set a fixed domain and then determine whether the keyword falls within it, but greenness cannot be defined by a specific series of words. Judgment Layer A is especially designed for the purpose mentioned above, undertaking the task of identifying green disclosure keywords.}

\re{We deploy a fine-tuned LLM ($LLM_A$) by prompt engineering to accomplish this task, and its judgment decison $\mathcal{J}$ can be given by the following mathematical form:}

\begin{align}\label{J}
\mathcal{J}=
\left\{
    \begin{aligned}
        &\:\; 1,\quad \mathcal{P}_{LLM_A,\:L}\left(G\right)\leq \mathcal{P}\left(\re{a_{i}}\mid G\right)\leq \mathcal{P}_{LLM_A,\:U}\left(G\right)
        \\ &\:\;0, \quad \mathcal{P}\left(\re{a_{i}}\mid G\right)> \mathcal{P}_{LLM_A,\:U}\left(G\right)\lor  \mathcal{P}\left(\re{a_{i}}\mid G\right)<\mathcal{P}_{LLM_A,\:L}\left(G\right)
    \end{aligned}\right.
\end{align}
Here $G$ represents the \re{preconfigured prompt template, which consists of  \texttt{Role Anchoring}, \texttt{Persona Setting}, \texttt{Task Description} and \texttt{Answer Template}. To enhance the transparency and reproducibility, we provide the format of our template in \autoref{templates}. Besides, it should be noted that \autoref{J} is only a mathematical expression of our judgment processing. In the experiment, we obtain the probability $\mathcal{P}\left(a_{i}\mid G\right)$ by pre-setting ``confidence'' in JSON schema in \texttt{Answer Template}, and the upper and lower bounds of the probability are implicit values of LLM, which are not ones must be calculated. What really works is $\mathcal{P}\left(a_{i}\mid G\right)$ — we will subsequently use its distribution as an important facter for selecting the most appropriate model.}

\re{The output of Judgment Layer A will be used as a dictionary. Annual report sentences containing keywords from this dictionary is regarded as the corporation's input sequence $\mathbf{S_2}$ for Judgment Layer B.}

\subsection{Effective Part of Green Disclosure}
\re{To accurately pinpoint corporations' substantive green disclosure, moving beyond mere keyword spotting, Judgment Layer B disentangles the symbolic disclosure from the efficitive improvements embedded within those keywords.} We \re{deploy} the same fine-tuned LLM expert for the detailed semantic analysis to determine the \re{symbolic and substantive} status. 

\re{Due to the main paths to improve LLM performance are iterative judgment \citep{madaan2023self, pang2024iterative} and enhanced retrieval \citep{lewis2020retrieval, salemi2025comparing}, we achieve the above two improvements in DeepGreen by setting the input text length and setting the Internet API tool. The most suitable plan will be selected based on the performance of ablation experiment.}

\re{Same as Judgment Layer A, the prompt template here also consists of \texttt{Role Anchoring}, \texttt{Persona Setting}, \texttt{Task Description} and \texttt{Answer Template}. Only \texttt{Task Description} is different from Judgment Layer A, adding detail guidelines for judgment based on the different categories of keywords identified in the previous layer. The major categories include file names, chemical substances or chemical terms, words that are directly related semantically to environmental protection, and other words that are commonly associated with corporations and factories. The specific content will also be provided in detail in \autoref{templates}.}

\re{The output of this layer is 0-1 judgment of the keywords in the corporation's annual report which are included in the green keyword dictionary. For the convenience of narration, the number of words judged as symbolical disclosure are referred as $Y$, and the number of words judged as substantive disclosure are referred as $X$. We refer to \cite{cao2022carbon, xu2023unveiling} and construct indicator Green Implement ($GI$) as:
\begin{align}
    GI = \dfrac{X}{X+Y}
\end{align}
$X+Y$ on the denominator represents the level of green disclosure in traditional research. The higher the value, the higher the level of green disclosure of the corporation, but the high value may be manipulated by inflating symbolic disclosure. Therefore, the significance of $GI$ lies in measuring the proportion of substantive disclosure in the overall green disclosure, effectively identifying the behaviour proposed by \cite{seele2017greenwashing} of greenwashing by piling up symbolic disclosure. Economically, when $GI \to 1$, the green disclosure is almost entirely backed by substantive improvements. Conversely, $GI \to 0$ reveals a greenwashing behaviour in which corporations rationally substitute symbolic keywords for substance, deferring compliance costs while still harvesting reputational benefits. $GI=0$ indicates that the corporation provides no relevant disclosure, or only piles up symbolic disclosure.}

\section{Experiment Details}\label{secExperiment}
\subsection{Dataset Source}

\re{In 2021, China issued new guidelines for the content and format of annual report information disclosure, placing ``environment and social responsibility'' separately in the fifth section and dividing it into two subsections: ``Environmental Information'' and ``Social Responsibility''. Therefore, we obtain all environmental information contexts in annual reports from 2021 to 2023 within seconds through simple regularization matching. To demonstrate DeepGreen's potential on large-scale dataset, our dataset includes A-share corporations that have published annual reports from 2021 to 2023 and avoids corporations labelled as $\mathrm{^*ST}$, $\mathrm{ST}$, $\mathrm{PT}$, ensuring the information integrity and comparability.}

\re{We excluded all financial corporations from the dataset not merely to follow standard research convention \citep{yuan2024exaggerating, chen2024mere, zhang2025encouraging}, but because the financial sector is institutionally unique. Financial institutions are simultaneously regulated by the central bank, banking-and-insurance regulator and securities regulator. There are significant differences in its internal operations and accounting systems compared to general industries. Moreover, the within-industry heterogeneity of financial corporations also exceeds the between-industry heterogeneity of non-financial corporations, rendering the two groups incomparable and necessitating an extra analytical framework for financial corporations.}

\re{In summary, the dataset includes 3123 corporations, of which 2498 corporations provide disclosure that meets the requirements, meaning that at least one observation in the three-year period has given a non-zero disclosure that needs to be processed. The corresponding indicator $GI$ of the remaining corporations will be directly assigned a value of 0. The descriptive statistics of observations that require processing by DeepGreen are presented in \autoref{data description}.}

\begin{table}[!htbp]
    \centering
    \caption{\centering Dataset Overview}
    \vspace{0.5em}
    \setlength{\tabcolsep}{2pt}
    \renewcommand{\arraystretch}{1.15}
    \begin{tabular}{l*8{p{1.5cm}}}
    \toprule
    & Obvs & Mean & Std & Min & 50\% & Max & Skew & Kurt \\
    \midrule
    Sentences\  & 7494 & 13.3487 & 21.4421 & 0 & 6 & 561 & 7.6765 & 124.7600 \\
    Words\  & 7494 & 2086.8989 & 3777.8241 & 0 & 571 & 83017 & 6.4582 & 78.6769 \\
    Target Words\  & 7494 & 407.9417 & 784.6850 & 0 & 97 & 18138 & 6.8974 & 88.5614 \\
    \bottomrule
    \end{tabular}
    \label{data description}
\end{table}

\subsection{Experiment Setting and Cost}
\re{Although DeepGreen is able to directly deliver $GI$ in a fully unsupervised manner, omitting any performance assessment may severely undermine the credibility of the resulting metrics and may consequently jeopardize the validity of our empirical studies. To guarantee reliability, we endow the two core judgment layers (A \& B) with explicit evaluation duties.}

\re{\textbf{Judgment Layer A} benchmarks the ability of different models to spot green-disclosure keywords under an identical prompt template. We run four parallel experiments: one locally-hosted LLaMA3-8B-Instruct \citep{dubey2024llama} and three API-called Chinese models Kimi-K2 \citep{team2025kimi}, Deepseek-V3.1 \citep{liu2024deepseek} and Qwen3-plus \citep{yang2025qwen3}. Each model receives the same prompt, and the extractions will be recorded and compared. \textbf{Judgment Layer B} performs an ablation experiment. The champion model emerging from Layer A is reused to contrast a control group (raw model) with two experimental groups: (i) a RAG-only condition, and (ii) a context-only condition, in which the model is allowed to refine its judgment solely through extended context. Considering cost-effectiveness, we will not investigate the method of using two improvement strategies simultaneously.}

\re{In terms of cost accounting, LLaMA3-8B-Instruct can be easily deployed on one RTX4080, while the computational cost of the three mainstream API-called products will be borne by the product side, which means that we only need to pay for the API fee. The total experimental cost is controlled within \$300. Meanwhile, in order to reduce the time cost, we use 100 concurrent requests so that LLM can give 100 judgment results simultaneously.}

\re{Across these experiments, we track performance indicators for classification tasks on random sampling verification. For Judgment Layer A, we randomly select 100 words, repeat ten times, and use the manual annotation of several students with financial backgrounds as the standard to calculate the various indicators for each sampling. Judgment Layer B repeats the procedure for 500 keyword–context pairs. The configuration that achieves the highest performance is ultimately entrusted with computing the indicator $GI$.}

\re{In binary classification, the four elementary outcomes are True Positive ($TP$), False Negative ($FN$), False Positive ($FP$) and True Negative ($TN$). Our evaluation indicators for both layers are accuracy (ACC), F1-score (F1), and Matthews Correlation Coefficient (MCC):
\begin{align}
    \mathrm{ACC} = \frac{TP + TN}{TP + TN + FP + FN}
\end{align}
\begin{align}
    \mathrm{F1} = \dfrac{2\ TP}{2\ TP+FN+FP}
\end{align}
\begin{align}
    \mathrm{MCC} = \frac{TP \cdot TN - FP \cdot FN}{\sqrt{(TP + FP)(TP + FN)(TN + FP)(TN + FN)}}
\end{align}
ACC measures the proportion of correctly classified instances among the total, but may be misleading under severe class imbalance. F1-score is the harmonic mean of precision and recall, providing a single scalar for class-specific performance. While MCC is a balanced measure even if classes are of very different sizes, which is regarded as one of the most reliable single metrics for binary and multiclass evaluation.}

\subsection{Results}
\subsubsection{Model Selecting from Layer A}
\begin{figure}[!htbp]
    \centering
    \caption{\centering Performance of Each Model in Identifying Green Disclosure Keywords}
    \includegraphics[width=0.9\linewidth]{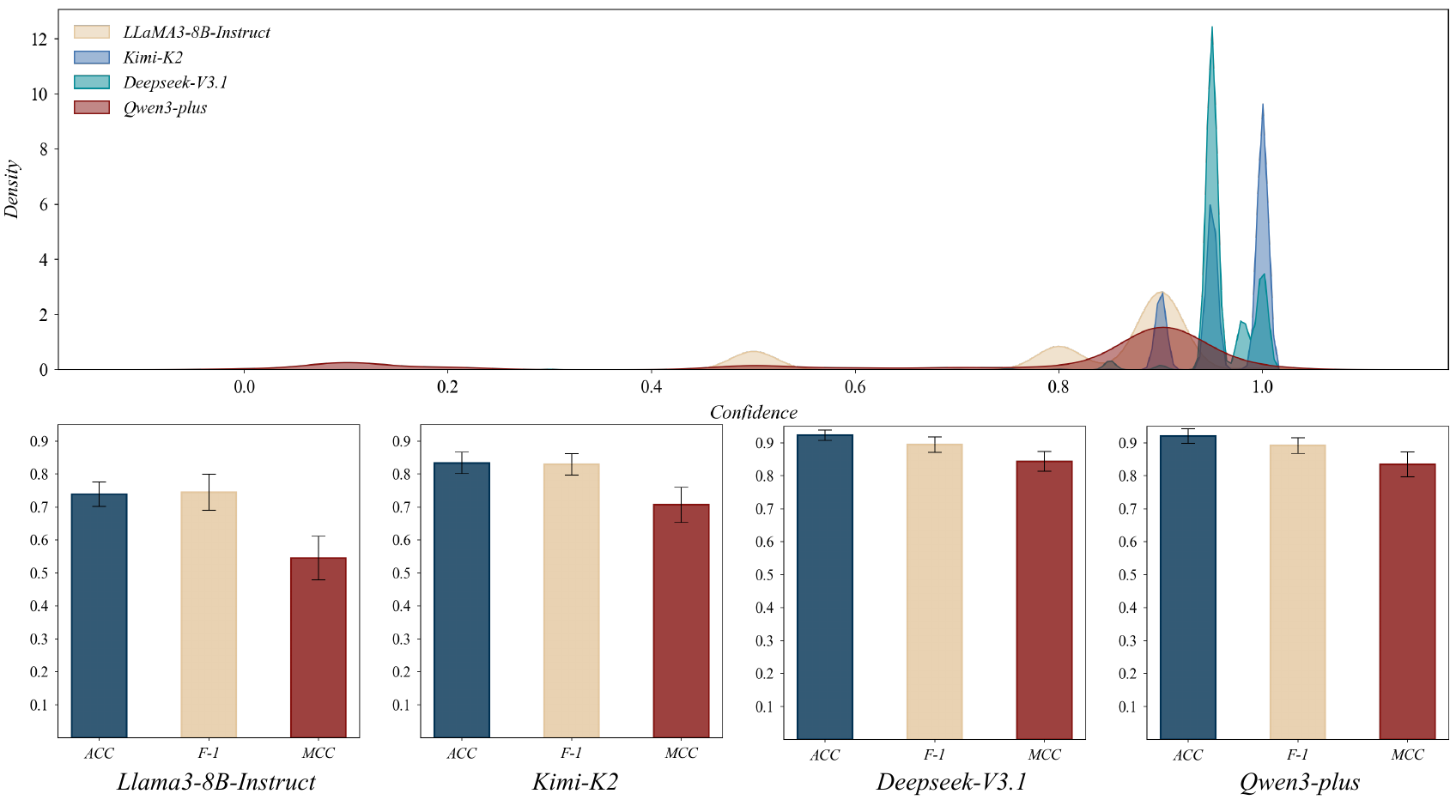}
    \label{layerA}
\end{figure}

\re{The discriminative performance of all experimental models in capturing green-disclosure keywords is synthesised in \autoref{layerA}. Simultaneously, we convert the judgment results' confidence level of each model into density maps to evaluate whether they are  randomly guessing.}

\re{The experiment results show that the performance of local LLaMA3-8B-Instruct falls short of optimal performance, with an ACC 10\% to 20\% lower than API-called models. Among the three API-called models, Kimi-K2 exhibites a relatively inferior performance. Although ACC and F1 only slip by 5$-$10 \%, MCC plummets by about 15 \%, a gap large enough to flag noisier classification. This larger decrement in MCC suggests that Kimi-K2's predictions contains a higher proportion of random guesses, indicating a less reliable classification behaviour.}

\re{At first glance, the remaining two contenders DeepSeek-V3.1 and Qwen3-plus present nearly identical point estimates (ACC, F1, MCC), but the probability density provides a negative answer. Qwen3-plus shows a markedly wider posterior, signalling that its high scores could still stem from sampling luck, and the broad spread reflects higher uncertainty in its judgments. As a consequence, we ultimately choose Deepseek-V3.1 as the champion model for the task of Judgment Layer B, for providing a very concentrated positive confidence while maintaining stable prediction performance.}

\subsubsection{Ablation Experiment in Layer B}
\begin{figure}[!htbp]
    \centering
    \caption{\centering Ablation Experiment}
    \label{Ablation}
    \includegraphics[width=0.8\linewidth]{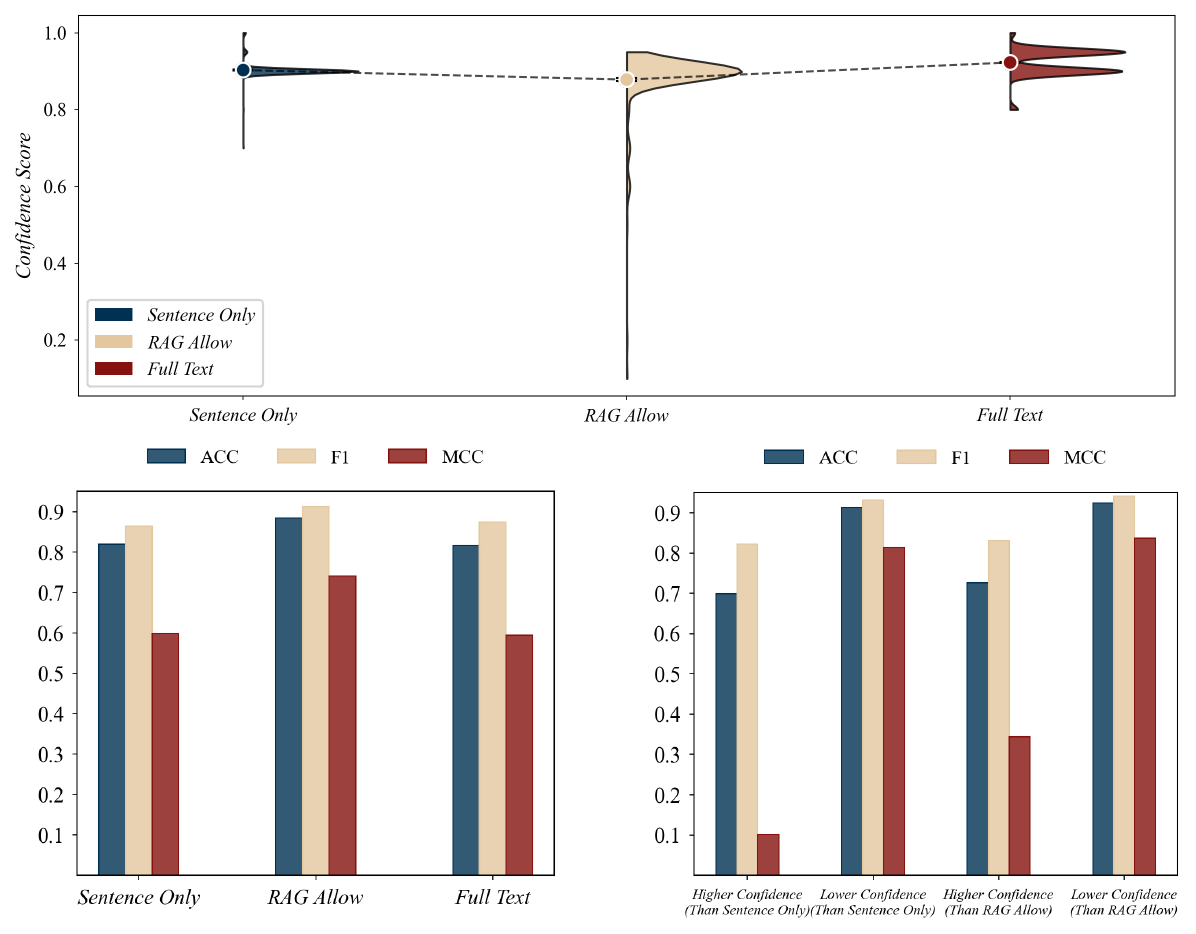}
\end{figure}

\re{Ablation experiment demonstrates that RAG yields higher performance in identifying substantive disclosure than merely extending context length. The detailed results are shown in three subgraphs in \autoref{Ablation}. The figure (lower left corner of \autoref{Ablation}) shows the performance comparison between the experimental group and the control group. It has been proven that the operation of expanding the context length has almost no substantial effect in this task, while the three indicators of the RAG method have significantly improved. Specifically, F1 reaches 90\%, indicating that the model has very few false positives and very few false negatives in identifying positive cases of substantive green disclosure.}

\re{The confidence distribution in the upper part allows us to conduct a deeper analysis of the model's ``black box''. We observe that, after the improvements, the model’s confidence distribution shifts. RAG renders the distribution slightly more dispersed yet still peaked around 0.9, while judgments of absolute certainty (>0.95) become markedly more frequent, mostly because the retrieved passages supply ample supplementary evidence. By contrast, extending the context converts the formerly unimodal distribution into a bimodal one, showing that the extra contextual tokens do exert a discernible effect.}

\re{However, when we stratify the extended-context model's outputs by confidence and benchmark their reliability (lower right corner of \autoref{Ablation}), we observe that the ``high-confidence'' predictions perform markedly worse, whereas the ``low-confidence'' predictions remain essentially unchanged, apparently reverting to the baseline judgments produced without context expansion. The evidence indicates that lengthening the context injects substantial invalid and redundant information, actively degrading the model's recognition performance.}

\re{Therefore, we choose the indicator $GI$ constructed by Deepseek-V3.1 with added RAG capability as corporate substantive green disclosure level.}

\section{Empirical Test}\label{secEmpirical}
\subsection{Hypothesis Development}
\re{We develop DeepGreen to validate LLMs’ capacity to detect greenwashing in firms’ symbolic disclosures and furthermore, to evaluate the empirical performance of its indicators.}

\re{Legitimacy-theory posits that corporations whose environmental performance lags behind societal expectations will seek symbolic means to restore legitimacy. When the gap between disclosed and actual environmental performance is wide, managers face intensified normative and coercive pressures to camouflage the shortfall through optimistic or selective disclosure \citep{magness2006strategic, brown1998public}. Such symbolic activities, however, do not reduce the underlying environmental risk, instead, they increase the probability that regulators will detect non-compliance during subsequent inspections. Because greenwashing signals a willingness to obscure information, it raises regulators’ belief that the corporation is withholding material violations. Empirically, therefore, greenwashing should covary positively with the likelihood of environmental penalty \citep{wu2020bad}, and we propose:}
\re{\begin{itemize}
    \item \noindent \textbf{H1:} \emph{\textbf{The adoption of corporate greenwashing behaviour presents a significantly positive correlation with being subject to environmental penalties.}}
\end{itemize}}

\re{While greenwashing increases regulatory exposure, the presence of green investors in management alters the corporation’s cost–benefit calculus in two complementary ways. First, environmentally focused investors possess the expertise to privately monitor the corporation’s environmental footprints, and the overstated environmental claims will be challenged from within the shareholders rather than only by external regulators \citep{jiang2021green}. Second, green investors supply not merely capital but also operational know-how—e.g., third-party verification bodies and cleaner-production technologies—that tangibly improve environmental performance \citep{liu2025does}. Therefore, we propose:}
\re{\begin{itemize}
    \item \noindent \textbf{H2:} \emph{\textbf{Green investors attenuate the positive association between corporate greenwashing and environmental penalties.}}
\end{itemize}}

\re{External audiences will continuously reassess a corporation’s environmental credibility by contrasting its public disclosures with observable attributes. When these attributes signal superior sustainability, they (including regulators) will form a more favorable prior belief about the corporation’s underlying environmental performance, thereby lowering the probability that symbolic exaggerations will trigger punitive enforcement \citep{rodrigue2013environmental, lyon2015means}. First, foreign owners typically originate from jurisdictions with stricter sustainability norms and oversight \citep{guo2021foreign}. Their equity stake signals that the corporation adheres to correspondingly higher standards, leading regulators to view identical greenwashing as less likely to reflect deliberate deceit and to impose fewer penalties. Second, measurable real actions, specifically documented energy-saving investments and third-party green certifications, serve as costly signals that the corporation has already incurred verifiable expenditures to improve environmental outcomes. These signals partially offset suspicions raised by overstated narratives \citep{plumlee2015voluntary}. Third, corporations with high governance quality exhibit stronger internal controls, more independent boards, and more transparent reporting systems. Regulators rationally expect such structures to limit managerial opportunism. Consequently, it will attenuate the positive greenwashing–penalty relation \citep{walker2012harm}. Finally, large corporation size confers visibility and reputational capital that magnify the downside of regulatory confrontation. These factors jointly reduce the likelihood that identified greenwashing escalates into formal penalties \citep{zhou2024threat, esposito2025sustainability}. Based on the above analysis, we further propose the hypothesis of heterogeneity:}
\re{\begin{itemize}
    \item \noindent \textbf{H3a:} \emph{\textbf{Foreign Investment and Ownership weaken the positive association between corporate greenwashing and environmental penalties.}}\\
    \item \noindent \textbf{H3b:} \emph{\textbf{Energy-saving activities weaken the positive association between corporate greenwashing and environmental penalties.}}\\
    \item \noindent \textbf{H3c:} \emph{\textbf{Environmental-friendly certification weakens the positive association between corporate greenwashing and environmental penalties.}}\\
    \item \noindent \textbf{H3d:} \emph{\textbf{High corporate governance quality weakens the positive association between corporate greenwashing and environmental penalties.}}\\
    \item \noindent \textbf{H3e:} \emph{\textbf{Large corporation size weakens the positive association between corporate greenwashing and environmental penalties.}}
\end{itemize}}

\subsection{Variable Description and Benchmark Model Setting}
\subsubsection{Dependent and Independent Variables}

\re{In the empirical stage, we use the corporate environmental penalties as the dependent variable. Following \cite{hu2023green}, we carefully match penalty‐decision announcements between corporations (and their subsidiaries) and the ecological environment bureaus, constructing the corporation environmental‐penalty variables \textbf{Vio} and \textbf{Vio\_num}. \textbf{Vio} is a binary dummy variable that equals 1 if the corporation (together with its subsidiaries) has received any form of penalty from the ecological-environment bureau during the focal year, and 0 otherwise. \textbf{Vio\_num} is a count variable that records the exact number of such penalties imposed on the corporation and its subsidiaries within the same year.}

\re{In constructing the independent variable, we note that DeepGreen focuses solely on the greenwashing behaviour defined by \cite{seele2017greenwashing}, and thus does not account for the selective disclosure issue highlighted by \cite{lyon2011greenwash} — namely, that annual reports may not fully reflect a corporation's actual environmental performance. Given this limitation, we adopt the approach proposed by \cite{yu2020greenwashing}, utilizing the environmental score from \textit{Huazheng} ESG ratings ($ESG_E$) as a proxy for corporate environmental performance. The reason for using \textit{Huazheng} rating here is that it clearly provides sub-scores in E, S, and G domains. Variable \textbf{Greenwashing} is thereby defined as a situation in which a corporation's disclosed environmental effectiveness exceeds the industry average, while its $ESG_E$ score remains below the industry average (\autoref{greenwashing}):
\begin{align}\label{greenwashing}
    Greenwashing = \left\{\begin{aligned}
        & 1, \quad GI>\overline{GI}\ \land\  ESG_E<\overline{ESG_E}\\
        & 0, \quad \mathrm{Others.}
    \end{aligned}\right.
\end{align}
When \textbf{Greenwashing} equals 1, the market receives a positive disclosure signal while remaining unaware that the corporation’s underlying environmental quality is below the industry benchmark. This phenomenon epitomizes the adverse-selection problem: the corporation selectively releases favorable green information, thereby capturing the reputational gains of a green image without incurring the attendant environmental costs.}

\re{Following the recent greenwashing literature \citep{sun2024bank,yuan2024exaggerating}, we employ a parsimonious set of firm-level controls: leverage (Lev; total liabilities divided by total assets), profitability (ROA; net income divided by total assets), growth prospects (Growth; revenue growth), ownership concentration (Top1; percentage stake of the largest shareholder), listing age (Listage; years since listing), capital intensity (Pfixa; log of net property, plant and equipment scaled by total assets), and operational efficiency (Psales; log value of total sales divided by total assets).}

\begin{figure}[!htbp]
    \centering
    \caption{\centering Correlation Analysis and VIF Test}
    \includegraphics[width=0.9\linewidth]{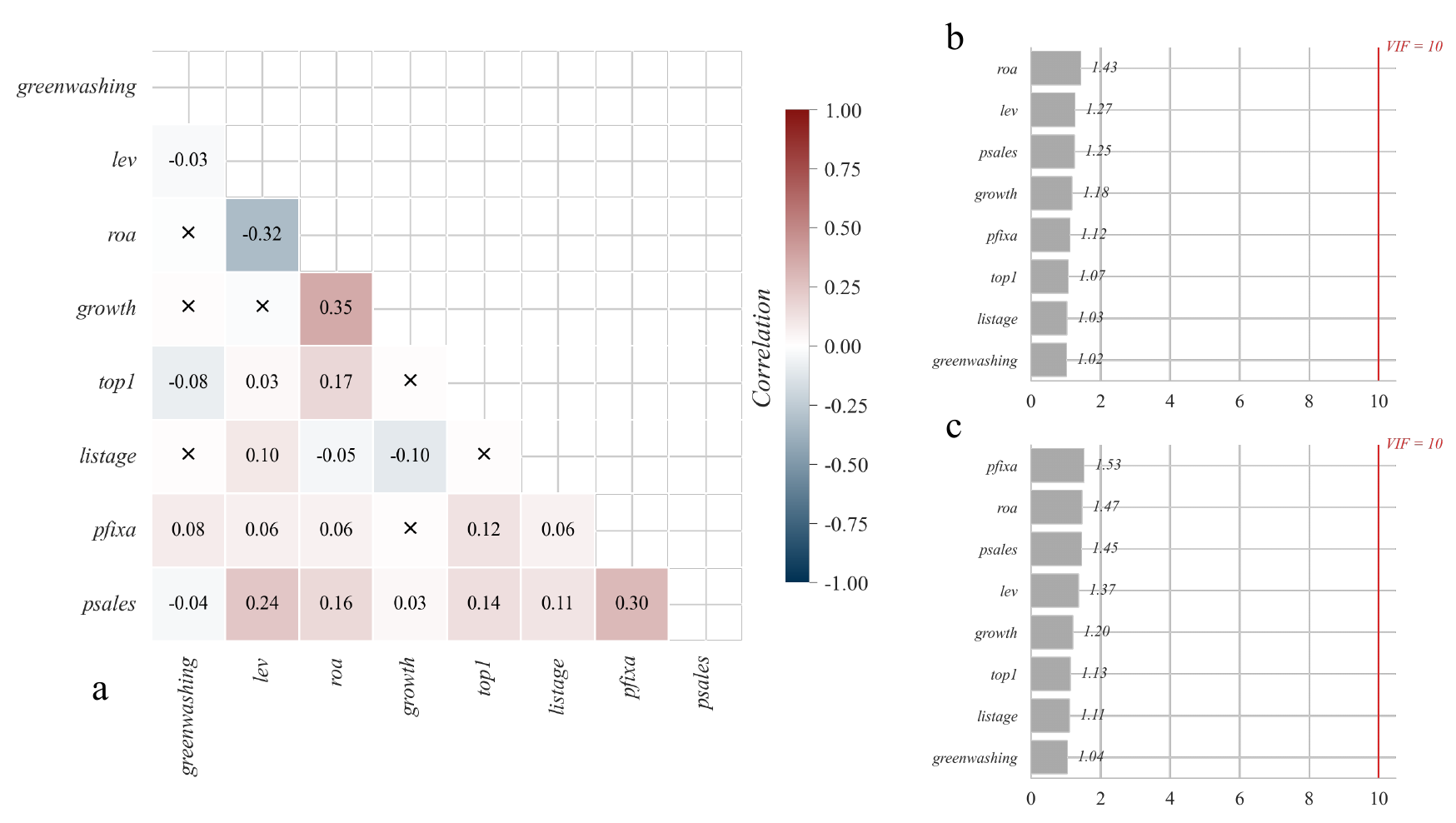}
    \label{vif}
\end{figure}

\re{We get the data of controls from Chinese Research Data Services (CNRDS) Platform. The descriptive statistics of all variables are reported in \autoref{description}, and the correlation matrix together with the multicollinearity diagnostics (VIF) are provided in \autoref{vif}. Subfigure \textbf{a} shows the correlation coefficient matrix between the independent variables, while \textbf{b} and \textbf{c} present the VIF results without controlling for industry and year fixed effects and after controlling for them. There is no significant multicollinearity observed in either case.}

\begin{table}[!htbp]
    \centering
    \caption{\centering Variable Descriptive Statistics}
    \vspace{0.5em}
    \label{description}
    \setlength{\tabcolsep}{2pt}
    \renewcommand{\arraystretch}{1.15}
    \begin{tabular}{l*8{p{1.5cm}}}
\toprule
 & Obvs & Mean & Std & Min & 50\% & Max & Skew & Kurt \\
\midrule
Vio & 9369 & 0.5171 & 0.4997 & 0.0000 & 1.0000 & 1.0000 & -0.0686 & -1.9957 \\
Vio\_num & 9369 & 0.5268 & 0.5184 & 0.0000 & 1.0000 & 2.0000 & 0.1019 & -1.4991 \\
\midrule
Greenwashing\ \  & 9369 & 0.2294 & 0.4205 & 0.0000 & 0.0000 & 1.0000 & 1.2876 & -0.3422 \\
Lev\ \  & 9369 & 0.4541 & 0.2008 & 0.0150 & 0.4489 & 1.9391 & 0.2809 & 0.0204 \\
ROA\ \  & 9369 & 0.0235 & 0.0768 & -1.1993 & 0.0281 & 0.7586 & -1.4920 & 17.2816 \\
Growth\ \  & 9369 & 0.0851 & 0.2292 & -0.9075 & 0.0524 & 5.6326 & 6.2406 & 98.2634 \\
Top1\ \  & 9369 & 0.3151 & 0.1443 & 0.0184 & 0.2923 & 0.8770 & 0.7126 & 0.1378 \\
Listage\ \  & 9369 & 23.4771 & 5.6836 & 6.0000 & 23.0000 & 69.0000 & 0.6799 & 2.7175 \\
Pfixa\ \  & 9369 & 12.7873 & 1.2273 & 5.7467 & 12.8099 & 19.6876 & -0.3760 & 2.4557 \\
Psales\ \  & 9369 & 14.1093 & 0.8684 & 10.1201 & 13.9998 & 18.8547 & 0.7510 & 1.5206 \\
\bottomrule
\end{tabular}
\end{table}

\subsubsection{Empirical Model}
\re{We employ a binary logistic regression model (Logit) to estimate the probability of corporate environmental violation occurrence as our benchmark. The model is specified as:
\begin{align}
    \ln\left(\dfrac{\mathcal{P}}{1-\mathcal{P}}\right)
= \beta_{0} + \beta_{1}Greenwashing + \mathbf{X}\boldsymbol{\gamma}+\gamma_{Year}+\gamma_{Ind},\quad \mathcal{P}=\mathcal{P}\left(Vio=1\mid Greenwashing,\ \mathbf{X}\right)
\end{align}
The model is estimated by maximum likelihood estimation (MLE), and here $\mathbf{X}$ represents the matrix of control variables mentioned above. $\gamma_{Year}$ and $\gamma_{Ind}$ respectively represents the fixed effects of year and industry. In subsequent robustness checks, we extend the models to address potential endogeneity and heterogeneous effects, which will be provided in section~\ref{Robustness}.}

\subsection{Benchmark Results}
\re{The benchmark employs the Logit model to examine the binary variable \textbf{Vio}. The analysis proceeds in a stepwise manner, beginning with univariate specifications and extending to multivariate settings, and throughout this process, fixed effects are incrementally introduced to control for unobserved heterogeneity. The detailed results are presented in \autoref{tabBenchmark}.}
\begin{table}[!htbp]
	\centering
	\caption{\centering Benchmark Results}
    \vspace{0.5em}
    \setlength{\tabcolsep}{2pt}
    \renewcommand{\arraystretch}{1.15}
	\begin{tabular}{l*6{>{\centering\arraybackslash}p{1.8cm}}}
		\toprule
		Vio & (1) & (2) & (3) & (4) & (5) & (6) \\
		\midrule
		Greenwashing & 0.2764*** & 0.3119*** & 0.2373*** & 0.2661*** & 0.2384*** & 0.2676*** \\
		& (0.0647) & (0.0658) & (0.0657) & (0.0663) & (0.0659) & (0.0665) \\
		Lev &  & 1.7903*** &  & 1.6253*** &  & 1.6372*** \\
		&  & (0.1593) &  & (0.1663) &  & (0.1668) \\
		ROA &  & 2.1991*** &  & 1.9436*** &  & 1.9782*** \\
		&  & (0.4186) &  & (0.4239) &  & (0.4260) \\
		Growth &  & 0.0056 &  & -0.0432 &  & -0.0381 \\
		&  & (0.1159) &  & (0.1111) &  & (0.1114) \\
		Top1 &  & 0.1842 &  & 0.0994 &  & 0.0983 \\
		&  & (0.1983) &  & (0.2072) &  & (0.2077) \\
		Listage &  & -0.0056 &  & -0.0053 &  & -0.0053 \\
		&  & (0.0050) &  & (0.0053) &  & (0.0053) \\
		Pfixa &  & 0.0543** &  & 0.0304 &  & 0.0307 \\
		&  & (0.0238) &  & (0.0275) &  & (0.0277) \\
		Psales &  & 0.0427 &  & 0.0555 &  & 0.0543 \\
		&  & (0.0356) &  & (0.0383) &  & (0.0384) \\
		\_cons & 0.0055 & -2.0893*** & 0.4005 & -1.5076** & 0.3453 & -1.5078** \\
		& (0.0313) & (0.5034) & (0.2598) & (0.6017) & (0.3455) & (0.6475) \\
        \midrule
		Year FE & NO & NO & YES & YES & YES & YES \\
		Ind FE & NO & NO & YES & YES & YES & YES \\
		Year \& Ind FE\ \ & NO & NO & NO & NO & YES & YES \\
		$N$ & 9369 & 9369 & 9369 & 9369 & 9366 & 9366 \\
		$R^{2} $& 0.0024 & 0.0264 & 0.0192 & 0.0357 & 0.0210 & 0.0377\\
		\bottomrule
	\end{tabular}\label{tabBenchmark}
    \vspace{0.5em}
    \noindent
	\begin{minipage}{0.84\textwidth}
		\small
        \vspace{0.5em}
		Standard errors in parentheses \\
		*p<0.1, **p<0.05, ***p<0.01
	\end{minipage}
\end{table}

\re{From the benchmark results of \autoref{tabBenchmark}, it is evident that the coefficient for the core independent variable \textbf{Greenwashing} (the corporate greenwashing behaviour) is positive and is significant at the 1\% level. This suggests a positive correlation between greenwashing behaviour and environmental penalities, supporting \textbf{H1}. To translate the statistical evidence into managerial intuition, we additionally report the average marginal effect (AME) of greenwashing on the probability of incurring an environmental penalty in \autoref{tabMargin}. This magnitude is stable whether we add controls or layer in year or industry fixed effects that AME remains tightly clustered — Per-unit increase of greenwashing raises the probability of incurring an environmental violation by roughly 6$\sim$7 percentage points, also indicating that about one out of every sixteen corporations that get environmental penalties has engaged in greenwashing behaviour. The constancy of the marginal effect underscores that the penalty premium attached to greenwashing is not an artefact of a particular set of covariates or unobserved heterogeneity, which is a robust feature of the benchmark for  supporting \textbf{H1}.} 

\begin{table}[!htbp]
	\centering
	\caption{\centering Marginal Effect of Benchmark}
    \vspace{0.5em}
    \setlength{\tabcolsep}{2pt}
    \renewcommand{\arraystretch}{1.15}
	\begin{tabular}{l*6{>{\centering\arraybackslash}p{1.8cm}}} 
		\toprule
		Vio      &  AME  &   AME &   AME &  AME &  AME &  AME                 \\
		\midrule
		Greenwashing        &      0.0634\textsuperscript{***}&      0.0751\textsuperscript{***}&      0.0577\textsuperscript{***}&      0.0632\textsuperscript{***}&      0.0578\textsuperscript{***}&      0.0634\textsuperscript{***}\\
                    &    (0.0157)         &    (0.0157)         &    (0.0159)         &    (0.0157)         &    (0.0159)         &    (0.0157)         \\
        Controls & NO & YES & NO & YES & NO & YES \\
        \midrule
        Year FE & NO & NO & YES & YES & YES & YES \\
		Ind FE & NO & NO & YES & YES & YES & YES \\
		Year \& Ind FE\ \ & NO & NO & NO & NO & YES & YES \\
		$N$   & 9369   & 9369   & 9369   & 9369   & 9369   & 9369                        \\
		\bottomrule
	\end{tabular}\label{tabMargin}
	\vspace{0.5em}
	\noindent
	\begin{minipage}{0.84\textwidth}
		\small
        \vspace{0.5em}
		Standard errors in parentheses \\
		*p<0.1, **p<0.05, ***p<0.01
	\end{minipage}
\end{table}

\subsection{Robustness Checks}\label{Robustness}
\subsubsection{Basic Robustness Test}
\re{Although the descriptive statistics reveal that no corporation in the sample receives more than two environmental penalties within a single year the dependent variable \textbf{Vio\_num} remains a count and the benchmark Logit model is no longer appropriate for this discrete nonnegative varible. Specifically, we deploy the OLS model, Poisson model, and Negative Binomial model. The model settings are as follows:
\begin{itemize}
    \item \text{OLS}
    \begin{align}
    \quad Vio\_num = \beta_0+\beta_1Greenwashing + \mathbf{X}\boldsymbol{\gamma}  +\gamma_{Year}+\gamma_{Ind} + \varepsilon
    \end{align}
    \item \text{Poisson \& Negative Binomial}
    \begin{align}
    \ln\Big[\mathbb{E}(\ Vio\_num\ )\Big] = \beta_0 &+ \beta_1 Greenwashing + \mathbf{X}\boldsymbol{\gamma} + \gamma_{Year} + \gamma_{Ind}
    \end{align}
\end{itemize}}

\re{The empirical results in \autoref{Alternative} indicate that whether we estimate the model with OLS, Poisson or Negative Binomial, the story is the same: greenwashing corporations are sanctioned more often. The magnitude is not merely statistically discernible but also practically relevant, implying that a one-unit increase in greenwashing is associated with an expected penalty count that lies about 8$\sim$16 percent higher, an effect that persists after the inclusion of year and industry fixed effects. Such uniformity across results alleviates concerns that the observed relation is an artefact of distributional assumptions or of the narrow support of the dependent variable. The behaviour of the control varibles reinforces this interpretation — After absorbing these sources of heterogeneity, the greenwashing coefficient remains virtually unchanged, indicating that the documented effect of greenwashing is not a surrogate for corporate financial performance.}

\begin{table}[!htbp]
	\centering
	\caption{\centering Robustness to Alternative Dependent Variables}
    \vspace{0.5em}
    \setlength{\tabcolsep}{2pt}
    \renewcommand{\arraystretch}{1.15}
	\begin{tabular}{l*6{>{\centering\arraybackslash}p{1.8cm}}} 
		\toprule
		  & (1)  & (2)  & (3)  & (4)  & (5)  & (6) \\
		  Vio\_num & OLS  & Possion  & Negative Binomial  & OLS  & Possion & Negative Binomial                          \\
		\midrule
		Greenwashing & 0.0877\textsuperscript{***} & 0.1627\textsuperscript{***}  & 0.1627\textsuperscript{***} & 0.0752\textsuperscript{***} & 0.1382\textsuperscript{***}  & 0.1382\textsuperscript{***}   \\
		~  & (0.0129)  & (0.0294)  & (0.0294)  & (0.0129)  & (0.0293)  & (0.0293) \\
		Lev  & 0.4460\textsuperscript{***} & 0.8602\textsuperscript{***}  & 0.8602\textsuperscript{***}  & 0.3991\textsuperscript{***} & 0.7759\textsuperscript{***}  & 0.7759\textsuperscript{***}   \\
		~   & (0.0302) & (0.0741)  & (0.0741)  & (0.0314)  & (0.0775)  & (0.0775) \\
		ROA   & 0.5325\textsuperscript{***} & 1.1298\textsuperscript{***}  & 1.1298\textsuperscript{***}  & 0.4639\textsuperscript{***} & 0.9766\textsuperscript{***}  & 0.9766\textsuperscript{***}   \\
		~   & (0.0839) & (0.2105)  & (0.2105)  & (0.0837)  & (0.2106)  & (0.2106)  \\
		Growth  & -0.0100   & -0.0216  & -0.0216   & -0.0233  & -0.0491    & -0.0491     \\
		~   & (0.0253) & (0.0513)  & (0.0513)  & (0.0243)  & (0.0512)  & (0.0512)   \\
		Top1  & 0.0512   & 0.0962   & 0.0962   & 0.0280  & 0.0526  & 0.0526    \\
		~   & (0.0374) & (0.0907)  & (0.0907)  & (0.0384)  & (0.0941)  & (0.0941)  \\
		Listage  & -0.0013    & -0.0025   & -0.0025    & -0.0011     & -0.0022    & -0.0022  \\
		~   & (0.0009) & (0.0023)  & (0.0023)  & (0.0010)  & (0.0024)  & (0.0024)   \\
		Pfixa    & 0.0129\textsuperscript{***} & 0.0252\textsuperscript{**}   & 0.0252\textsuperscript{**}   & 0.0073 & 0.0149  & 0.0149  \\
		~   & (0.0045) & (0.0110)  & (0.0110)  & (0.0052)  & (0.0129)  & (0.0129)  \\
		Psales   & 0.0117\textsuperscript{*}   & 0.0198   & 0.0198  & 0.0147\textsuperscript{**}  & 0.0258   & 0.0258 \\
		~   & (0.0069) & (0.0163)  & (0.0163)  & (0.0073)  & (0.0177)  & (0.0177)   \\
		\_cons  & -0.0234  & -1.6847\textsuperscript{***} & -1.6847\textsuperscript{***} & 0.0354 & -1.4208\textsuperscript{***} & -1.4208\textsuperscript{***}  \\
		~   & (0.0958) & (0.2302)  & (0.2302)  & (0.1039)  & (0.2727)  & (0.2727)  \\
        \midrule
		Year FE  &      &     &        & YES    & YES    & YES    \\
		Ind FE  &      &     &        & YES    & YES    & YES   \\
		lnalpha  & ~  & ~   & -22.0151  & ~   & ~    & -22.0151   \\
		\textit{N}  & 9369   & 9369    & 9369    & 9369    & 9369   & 9369  \\
		$R^2$  & 0.0367     & ~       & ~     & 0.0495     & ~      & ~     \\
		$R^2\_p$    & ~    & 0.0108   & 0.0108   & ~   & 0.0147  & 0.0147   \\
		\bottomrule
	\end{tabular}\label{Alternative}
	\vspace{0.5em}
	\noindent 
	\begin{minipage}{0.83\textwidth}
		\small
        \vspace{0.5em}
		Standard errors in parentheses \\
		*p<0.1, **p<0.05, ***p<0.01
	\end{minipage}
\end{table}

\subsubsection{Instrumental-Variable Check}

\begin{table}[!htbp]
\centering
\caption{\centering IV Results}
\vspace{0.5em}
\setlength{\tabcolsep}{2pt}
\renewcommand{\arraystretch}{1.15}
\begin{tabular}{l*4{>{\centering\arraybackslash}p{1.8cm}}}
\toprule
& (1) & (2) & (3) & (4) \\
\cmidrule(l){2-5}
IV & \multicolumn{2}{c}{IV} & \multicolumn{2}{c}{Dynamic IV} \\
\midrule
\multicolumn{5}{l}{\textbf{Stage One: Greenwashing}} \\
\midrule
Greenwashing\_lag & 0.7949*** & 0.7949*** & 0.7949*** & 0.7949*** \\
& (0.0102) & (0.0102) & (0.0102) & (0.0102) \\
\_cons & 0.0286*** & 0.0286*** & 0.0286*** & 0.0286*** \\
& (0.0024) & (0.0024) & (0.0024) & (0.0024) \\
\midrule
\multicolumn{5}{l}{\textbf{Stage Two: Vio}} \\
\midrule
Greenwashing & 0.0790*** & 0.0829*** & 0.0475*** & 0.0507*** \\
& (0.0217) & (0.0214) & (0.0161) & (0.0162) \\
Lev & & 0.3846*** & & 0.2100*** \\
& & (0.0417) & & (0.0324) \\
ROA & & 0.5618*** & & 0.3736*** \\
& & (0.1056) & & (0.0872) \\
Growth & & -0.0523* & & -0.0215 \\
& & (0.0284) & & (0.0271) \\
Top1 & & 0.0251 & & 0.0122 \\
& & (0.0539) & & (0.0396) \\
Listage & & -0.0013 & & -0.0009 \\
& & (0.0014) & & (0.0010) \\
Pfixa & & 0.0094 & & 0.0070 \\
& & (0.0070) & & (0.0052) \\
Psales & & 0.0099 & & 0.0032 \\
& & (0.0098) & & (0.0073) \\
Vio\_lag & &  & 0.4285*** & 0.4155*** \\
& &  &(0.0126) & (0.0128) \\
\_cons & 0.6060*** & 0.1674 & 0.3566*** & 0.1367 \\
& (0.0694) & (0.1559) & (0.0527) & (0.1152) \\
\midrule
lnsig\_1\_cons & -1.4383*** & -1.4383*** & -1.4383*** & -1.4383*** \\
& (0.0234) & (0.0234) & (0.0234) & (0.0234) \\
lnsig\_2\_cons & -0.7068*** & -0.7183*** & -0.8086*** & -0.8129*** \\
& (0.0022) & (0.0031) & (0.0068) & (0.0069) \\
atanhrho\_12\_cons & -0.0279* & -0.0251* & -0.0112 & -0.0100 \\
& (0.0152) & (0.0152) & (0.0163) & (0.0163) \\
$N$ & 6246 & 6246 & 6246 & 6246 \\
\bottomrule
\end{tabular}\label{tabIV}
\vspace{0.5em}
\noindent 
\begin{minipage}{0.63\textwidth}
		\small
        \vspace{0.5em}
		Standard errors in parentheses \\
		*p<0.1, **p<0.05, ***p<0.01
	\end{minipage}
\end{table}

\re{To mitigate the potential endogeneity of the model, we instrument current greenwashing with its once-lagged value (\textbf{Greenwashing\_lag}) in \autoref{tabIV}. The first-stage F-statistic far exceeds the rule-of-thumb threshold of ten, virtually eliminating weak-instrument concerns. Besides, the estimated elasticity of sanctions to greenwashing is about 0.05$\sim$0.08, which remains unchanged after comprehensive financial controls are added, indicating that the penalty is not a proxy for opaque or fragile corporations but a direct response to violation. Once a corporation adopts greenwashing behaviour, its probability of being subject to environmental penalties will significantly increase.}

\re{Columns (3) and (4) add the lagged penalty (\textbf{Vio\_lag}) to capture violation inertia. The coefficient on this control is significantly positive, confirming that previous violations are more likely to reoffend. The coefficient of \textbf{Greenwashing} in the second stage stays positive and significant at the same magnitude,showing that even among corporations with identical violation histories those that embellish or greenwash are punished more often. Thus greenwashing is not merely correlated with violation, but also an independent precursor of future infractions.}

\re{The ancillary parameter $\mathrm{atanhrho\_{12}\_cons}$ quantifies the residual correlation between the first-stage and second-stage errors. When \textbf{Vio\_lag} is excluded the estimate is marginally significant, suggesting a modest common unobserved component. Once \textbf{Vio\_lag} is added, the correlation falls to $-0.011$ and loses significance, indicating that past enforcement captures the latent corporate heterogeneity that previously leaked into both stages. Because disclosure rhetoric is sticky yet enforcement reacts to contemporaneous signals, lagged greenwashing affects new sanctions only through the persistent component that survives last year’s investigation. Hence, \textbf{Greenwashing\_lag} satisfies the exclusion restriction: any residual impact on current sanctions operates solely through the persistent rhetorical distortion it induces.}

\subsubsection{Propensity Score Matching (PSM) Estimates}
\re{To address the potential selection bias in our analysis, we employ Propensity Score Matching (PSM). By matching treated and control units based on their propensity scores, we create a more balanced sample, thereby enhancing the credibility of our causal inferences. This approach is particularly relevant in our study, as it allows us to better isolate the effect of greenwashing on corporate penalties while controlling for other confounding factors.}

\begin{figure}[!htbp]
    \centering
    \caption{\centering PSM Variable Matching Results}
    \includegraphics[width=0.6\linewidth]{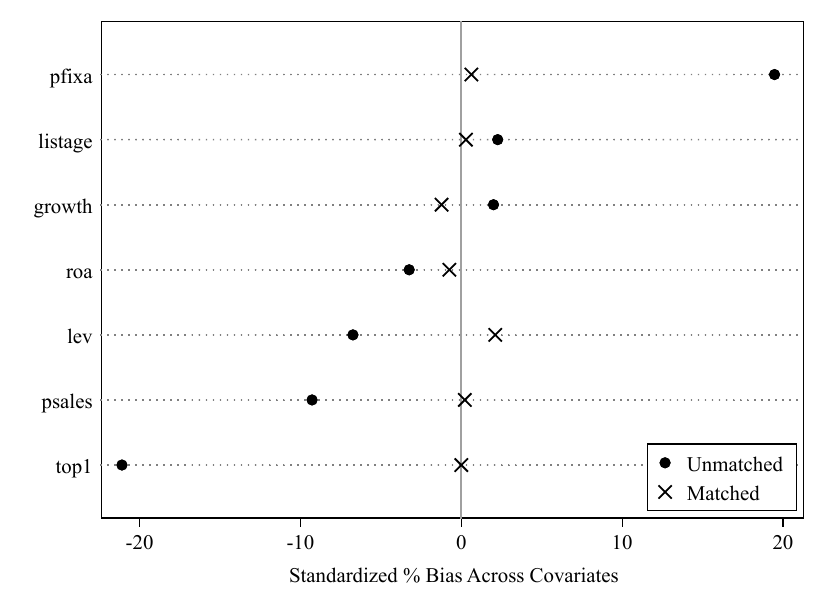}
    \label{PSM}
\end{figure}

\re{The matching is performed using the nearest neighbor matching (1:1 matching) to ensure that only units with sufficiently close propensity scores are matched. The propensity score matching results are illustrated in \autoref{PSM}. These results show that the bias has been significantly reduced for all covariates, indicating that the matching process has effectively balanced the treatment and control groups. Specifically, the standardized \% bias for variables such as listage growth, ROA, leverage, and sales has been brought close to zero, suggesting that the matched groups are now comparable in terms of these key characteristics. This successful matching ensures that any subsequent analysis of the treatment effect is less likely to be confounded by differences in these covariates.}

\re{After ensuring the balance of covariates through PSM, we employ both Logit and Probit models to estimate the treatment effect. The results of these regressions are presented in \autoref{tabPSM}. Both the Logit and Probit models consistently show that greenwashing is significantly associated with an increase in corporate penalties, further reinforcing our initial findings. The coefficients on greenwashing remain statistically significant across both models, indicating that the effect of greenwashing on penalties is robust to the choice of estimation technique and potential selection bias.}

\begin{table}[!htbp]
	\centering
	\caption{\centering PSM Regression Results}
    \vspace{0.5em}
    \setlength{\tabcolsep}{2pt}
    \renewcommand{\arraystretch}{1.15}
	\begin{tabular}{l*4{>{\centering\arraybackslash}p{1.8cm}}}
		\toprule
		& (1) & (2) & (3) & (4) \\
        \cmidrule(r){2-3} \cmidrule(l){4-5}
		Vio & \multicolumn{2}{c}{Logit} & \multicolumn{2}{c}{Probit} \\
		\midrule
		greenwashing & 0.3011*** & 0.3312*** & 0.1878*** & 0.2008***  \\ 
        ~ & (0.0756) & (0.0769) & (0.0470) & (0.0473)  \\ 
        size & ~ & 0.0155*** & ~ & 0.0074***  \\ 
        ~ & ~ & (0.0056) & ~ & (0.0024)  \\ 
        lev & ~ & 1.2851*** & ~ & 0.8132***  \\
        ~ & ~ & (0.2402) & ~ & (0.1426)  \\
        roa & ~ & 0.9795* & ~ & 0.6393*  \\ 
        ~ & ~ & (0.5679) & ~ & (0.3474)  \\ 
        growth & ~ & -0.1316 & ~ & -0.0790  \\ 
        ~ & ~ & (0.1386) & ~ & (0.0871)  \\ 
        top1 & ~ & -0.4449 & ~ & -0.2259  \\ 
        ~ & ~ & (0.3162) & ~ & (0.1932)  \\ 
        listage & ~ & -0.0139* & ~ & -0.0084*  \\
        ~ & ~ & (0.0074) & ~ & (0.0045)  \\ 
        pfixa & ~ & -0.1013** & ~ & -0.0587**  \\ 
        ~ & ~ & (0.0488) & ~ & (0.0294)  \\ 
        psales & ~ & 0.0219 & ~ & 0.0192  \\
        ~ & ~ & (0.0628) & ~ & (0.0382)  \\ 
        \_cons & 0.1520 & 0.8621 & 0.0948 & 0.3893  \\ 
        ~ & (0.3517) & (0.9582) & (0.2203) & (0.5966)  \\ 
        \midrule
        Year FE & YES & YES  & YES  & YES   \\
		Ind FE & YES  & YES  & YES  & YES   \\
        $N$ & 4298 & 4298 & 4298 & 4298  \\ 
        $R^2\_p$ & 0.0202 & 0.0441 & 0.0203 & 0.0429\\
		\bottomrule
	\end{tabular}\label{tabPSM}
    \vspace{0.5em}
    \noindent 
    \begin{minipage}{0.59\textwidth}
		\small
        \vspace{0.5em}
		Standard errors in parentheses \\
		*p<0.1, **p<0.05, ***p<0.01
	\end{minipage}
\end{table}

\subsubsection{Placebo Test}
\begin{figure}[!htbp]
    \centering
    \caption{\centering Placebo Test Results}
    \includegraphics[width=0.8\linewidth]{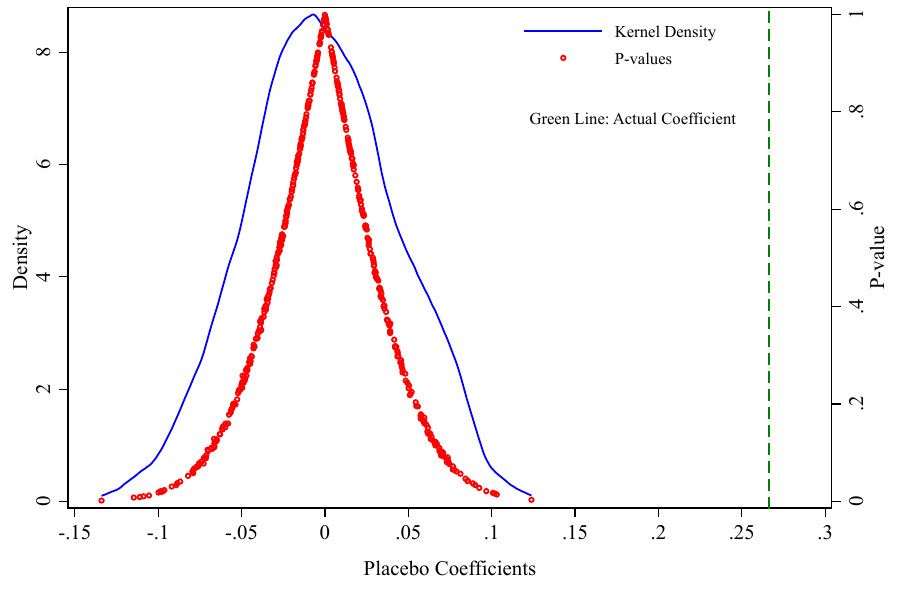}
    \label{Placebo}
\end{figure}

\re{To further validate the robustness of our findings, we conduct a placebo test. In this test, we replace the greenwashing variable with a randomly generated placebo variable that has no theoretical basis for affecting corporate penalties. If the placebo variable were to yield statistically significant results, it would suggest that our initial findings might be driven by spurious correlations rather than a true causal effect. The results of the placebo test are presented by a kernel density plot in \autoref{Placebo}. The green line represents the coefficient of the actual green washing variable, while the distribution of the placebo coefficient is concentrated around zero and significantly different from the actual coefficient. This indicates that our original results were not caused by random noise or model errors. This placebo test provides additional evidence that the observed relationship between greenwashing and corporate penalties is robust and reliable, strongly supports \textbf{H1}.}

\subsection{Moderating Effects}

\begin{table}[!htbp]
	\centering
	\caption{\centering Moderating Effects}
    \vspace{0.5em}
    \setlength{\tabcolsep}{2pt}
    \renewcommand{\arraystretch}{1.15}
	\begin{tabular}{l*4{>{\centering\arraybackslash}p{1.8cm}}}
		\toprule
		& (1) & (2) & (3) & (4) \\ 
		\cmidrule(r){2-3} \cmidrule(l){4-5}
		Vio & \multicolumn{2}{c}{Logit} & \multicolumn{2}{c}{Probit}  \\ 
        \midrule
		Greenwashing & 0.4095*** & 0.3826*** & 0.2533*** & 0.2368*** \\
		& (0.0779) & (0.0755) & (0.0483) & (0.0468) \\
		ESG Investor & 0.3953*** &  & 0.2472*** &  \\
		& (0.0629) &  & (0.0390) &  \\
		Greenwashing $\times$ ESG Investor & -0.3939*** &  & -0.2437*** &  \\
		& (0.1236) &  & (0.0768) &  \\
        Base &  & 0.2808*** &  & 0.1747*** \\
		&  & (0.0418) &  & (0.0256) \\
		Greenwashing $\times$ Base &  & -0.2153** &  & -0.1319** \\
		&  & (0.0869) &  & (0.0539) \\
		Lev & 1.5606*** & 1.5385*** & 0.9573*** & 0.9444*** \\
		& (0.1652) & (0.1651) & (0.1001) & (0.1001) \\
		ROA & 1.3793*** & 1.2032*** & 0.8356*** & 0.7290*** \\
		& (0.4237) & (0.4235) & (0.2587) & (0.2590) \\
		Growth & -0.0949 & -0.1391 & -0.0567 & -0.0852 \\
		& (0.1069) & (0.1065) & (0.0667) & (0.0666) \\
		Top1 & 0.0813 & 0.1154 & 0.0551 & 0.0759 \\
		& (0.2073) & (0.2073) & (0.1281) & (0.1280) \\
		Listage & -0.0040 & -0.0039 & -0.0025 & -0.0025 \\
		& (0.0053) & (0.0053) & (0.0033) & (0.0033) \\
		Pfixa & 0.0266 & 0.0243 & 0.0163 & 0.0148 \\
		& (0.0275) & (0.0276) & (0.0170) & (0.0170) \\
		Psales & 0.0429 & 0.0380 & 0.0272 & 0.0242 \\
		& (0.0382) & (0.0383) & (0.0236) & (0.0236) \\
		\_cons & -1.3928** & -1.2938** & -0.8654** & -0.8022** \\
		& (0.5979) & (0.5994) & (0.3688) & (0.3696) \\
        \midrule
		Year FE & YES & YES  & YES  & YES   \\
		Ind FE & YES  & YES  & YES  & YES   \\
		$N$ & 9369 & 9369 & 9369 & 9369 \\
		$R^{2} \_p$ & 0.0400 & 0.0411 & 0.0399 & 0.0411 \\
        \bottomrule
	\end{tabular}\label{tabModerating}
    \vspace{0.5em}
    \noindent 
    \begin{minipage}{0.74\textwidth}
		\small
        \vspace{0.5em}
		Standard errors in parentheses \\
		*p<0.1, **p<0.05, ***p<0.01
	\end{minipage}
\end{table}

\re{With the purpose of testing \textbf{H2}, which posits that green investors can mitigate the positive association between corporate greenwashing and environmental penalties, we examine two moderating variables: (1) \textbf{ESG Investor}, a binary indicator reflecting the presence of green investors in a corporation’s ownership structure; (2) \textbf{Base}, a continuous measure represented by log of the number of green investors (plus one to avoid taking the log of zero). The results are presented in \autoref{tabModerating}. The empirical findings offer compelling evidence underscoring the salient moderating function of green investors. Both the mere presence of green investors and the base of them exert a dampening effect on the propensity for greenwashing to escalate environmental penalties. Nevertheless, another noteworthy finding is the positive direct correlation observed between green investors and environmental penalties.}

\re{First, both moderators significantly attenuate the positive relationship between greenwashing and environmental penalties. It is evidenced by the negative and statistically significant coefficients on the interaction terms $\mathrm{Greenwashing \times ESG Investor}$ and $\mathrm{Greenwashing \times Base}$ across all model specifications, supporting \textbf{H2} by suggesting that green investors reduce the likelihood that greenwashing behaviour leads to environmental violations.}

\re{Second,  we observe that the direct coefficient of \textbf{ESG Investor} is positive and significant, implying that on average, corporations with green investors are more likely to face environmental penalties. This seemingly counterintuitive result can be explained through an institutional or scrutiny-based lens, that corporations attracting green investors often operate in environmentally sensitive sectors or face greater regulatory and public scrutiny. Thus, even if these investors curb the penalty-enhancing effect of greenwashing, their very presence may signal higher baseline exposure to environmental enforcement. In other words, while green investors help moderate the adverse outcome of greenwashing, they do not eliminate the underlying regulatory risks associated with the corporation’s industrial background.}

\subsection{Heterogeneity Analysis}
\re{This section examines potential heterogeneity through regression analysis of sub samples, and the results are summarized in \autoref{tabHeterogeneity}. Overall, there is no significant positive relationship between greenwashing behaviour and environmental violations within the group of foreign investment or ownership corporations, energy-saving corporations, obtained environmental certifications, and large-scale corporations. However, the level of corporate governance has a relatively small impact on the positive relationship between greenwashing behaviour and environmental violations. This results support \textbf{H3a}, \textbf{H3b}, \textbf{H3c}, and \textbf{H3e}.}

\re{Foreign ownership may have a mitigating effect on greenwashing behaviour, potentially due to stricter regulatory environments and higher standards in foreign jurisdictions. Similarly, corporations engaged in energy-saving activities are less likely to exhibit greenwashing behaviour, possibly because they are more focused on genuine environmental stewardship. Obtaining environmental certifications may also reduce the likelihood of greenwashing, as these certifications often require adherence to strict environmental standards. In contrast, large corporations, compared to small and medium-scale corporations, have stronger market reputations and more stable market positions, thereby unless they are found to have serious violations, it is less likely that they will be accused of greenwashing. On the other hand, regulators are not very sensitive to the factor of measuring corporate governance level through ratings.Taking into account these five types of heterogeneity, regulators are relatively more inclined to focus on and adhere to fact-based judgments. In a market where greenwashing exists, regulators tend to see more of the ``actions'' rather than the ``words'', and even less so the consistency between words and actions. As a consequence, this approach may inadvertently allow some greenwashing behaviour to go unchecked.}

\begin{sidewaystable}[!htbp]
    \centering
    \caption{\centering Heterogeneity Analysis Results}
    \vspace{0.5em}
    \setlength{\tabcolsep}{2pt}
    \renewcommand{\arraystretch}{1.15}
    \begin{tabular}{l*{11}{>{\centering\arraybackslash}p{1.7cm}}}
		\toprule
		& \multicolumn{2}{c}{Foreign Investment/Ownership} & \multicolumn{2}{c}{Energy Saving} & \multicolumn{2}{c}{Environmental Certification} & \multicolumn{2}{c}{Governance Level} & \multicolumn{3}{c}{Corporate Scale} \\
		\cmidrule(r){2-3} \cmidrule(l){4-5}   \cmidrule(l){6-7}  \cmidrule(l){8-9}  \cmidrule(l){10-12}
		& NO & YES & NO & YES & NO & YES & Weak & Strong & Small & Medium & Large \\
        \midrule
		Greenwashing & 0.2965*** & -0.4288 & 0.3166*** & 0.2001 & 0.3045*** & 0.2272 & 0.2904*** & 0.2635*** & 0.3615*** & 0.2859** & -0.6125 \\
		& (0.0682) & (0.2912) & (0.0717) & (0.1465) & (0.0696) & (0.1866) & (0.0921) & (0.0874) & (0.0760) & (0.1399) & (0.5656) \\
		Lev & 1.6209*** & 1.8607** & 1.3701*** & 2.4219*** & 1.4454*** & 2.5763*** & 1.1266*** & 2.6143*** & 1.0690*** & 1.3095*** & -1.2105 \\
		& (0.1708) & (0.7590) & (0.1786) & (0.3954) & (0.1730) & (0.4672) & (0.2195) & (0.2415) & (0.1872) & (0.4112) & (1.8790) \\
		ROA & 1.9981*** & 0.9353 & 1.5291*** & 2.5780** & 1.6997*** & 2.7761** & 1.4308*** & 2.1481*** & 0.8730* & 2.6608** & -6.7659 \\
		& (0.4377) & (1.8044) & (0.4542) & (1.0599) & (0.4402) & (1.2253) & (0.5486) & (0.5990) & (0.4483) & (1.1649) & (4.6681) \\
		Growth & -0.0069 & -0.5840 & -0.0168 & -0.1003 & -0.0556 & 0.0767 & 0.1378 & -0.3081** & -0.0158 & -0.3194 & 1.8862 \\
		& (0.1153) & (0.5124) & (0.1179) & (0.3256) & (0.1145) & (0.3958) & (0.1825) & (0.1486) & (0.1179) & (0.2487) & (1.3251) \\
		Top1 & 0.0403 & 0.4459 & 0.0738 & -0.2989 & 0.0611 & -0.2534 & -0.0118 & 0.0609 & -0.0848 & -0.7456** & 1.6720 \\
		& (0.2189) & (0.7337) & (0.2379) & (0.3855) & (0.2251) & (0.4476) & (0.3068) & (0.2593) & (0.2677) & (0.3615) & (1.3803) \\
		Listage & -0.0037 & -0.0419* & -0.0066 & -0.0096 & -0.0035 & -0.0214* & -0.0116 & -0.0002 & -0.0050 & -0.0201* & -0.0004 \\
		& (0.0054) & (0.0242) & (0.0060) & (0.0098) & (0.0057) & (0.0113) & (0.0073) & (0.0067) & (0.0064) & (0.0105) & (0.0261) \\
		Pfixa & 0.0271 & 0.1422 & 0.0254 & 0.0272 & 0.0174 & 0.1354 & 0.0425 & 0.0390 & 0.0126 & -0.0626 & 0.1768 \\
		& (0.0287) & (0.1093) & (0.0293) & (0.0609) & (0.0283) & (0.0839) & (0.0370) & (0.0375) & (0.0323) & (0.0554) & (0.2320) \\
		Psales & 0.0677* & -0.1532 & 0.0695 & -0.0689 & 0.0498 & -0.0422 & 0.0749 & 0.0026 & 0.0162 & -0.1307* & -0.5103** \\
		& (0.0395) & (0.1827) & (0.0423) & (0.0803) & (0.0404) & (0.1021) & (0.0519) & (0.0524) & (0.0468) & (0.0711) & (0.2496) \\
		\_cons & -1.6633*** & 0.5386 & -1.5792** & 0.5505 & -1.2770** & -0.1231 & -1.7466** & -1.2678 & -0.5623 & 3.4095*** & 6.9128** \\
		& (0.6144) & (2.7172) & (0.6772) & (1.1367) & (0.6428) & (1.6995) & (0.8399) & (0.7934) & (0.7595) & (1.1637) & (3.3910) \\
        \midrule
		Year & Control & Control  & Control & Control& Control & Control  & Control & Control& Control & Control  & Control \\
		Ind  & Control   & Control  & Control    & Control  & Control & Control  & Control & Control& Control & Control  & Control  \\
		$N$ & 8814 & 543 & 7160 & 2205 & 7790 & 1576 & 4028 & 5340 & 6312 & 2679 & 374 \\
		$R^{2} \_p$ & 0.0367 & 0.0614 & 0.0319 & 0.0535 & 0.0336 & 0.0561 & 0.0338 & 0.0475 & 0.0227 & 0.0489 & 0.1323 \\
        \bottomrule
	\end{tabular}\label{tabHeterogeneity}
	\vspace{0.5em}
	\noindent
	\begin{minipage}{\linewidth}
		\small
        \vspace{0.5em}
		Standard errors in parentheses \\
		*p<0.1, **p<0.05, ***p<0.01
	\end{minipage}
\end{sidewaystable}

\section{Discussion and Conclusion}\label{secConclusion}

\re{The paper aims to address two fundamental challenges in greenwashing studies. First, we evaluate whether LLMs can effectively distinguish between substantive and symbolic environmental disclosures, thereby mitigating greenwashing that arises from the accumulation of symbolic statements. Second, we examine whether combining LLM-Driven disclosure measures with ESG performance can reduce bias of detecting greenwashing from selective disclosure. The experimental results confirm the first question, that DeepGreen is able to differentiates between these two types of disclosures. The empirical analysis replies the second question, demonstrating that corporations identified as greenwashing through face a significantly higher probability of environmental penalties, establishing a causal link rather than mere correlation.}

\re{However, there are lots of studies continue to concern the reliability of unsupervised text analysis, and the concern is often framed as a demand for error-free from LLMs. Such a demand is unrealistic, because when human themselves label subtle constructs such as greenwashing, the annotators' agreement will also rarely reach perfection. In the present study, the best performing LLM achieved an F-1 score that lies within the range typically reported for trained human analysts who work on the same task. When the prompt is specific and the evaluation rubric is transparent, the model behaves as a consistent research assistant rather than as an opaque black box. The ablation exercise further shows that retrieval augmented generation raises precision by grounding the model in external evidence, a finding that directly releaves the hallucination problem. Three incremental strategies can strengthen trust even more: (1) Archive retrieved documents and pool validated passages into an open knowledge base. (2) Deploy several LLMs in parallel, log their discussions or independent expert-judgments. (3) Train expert LLM through methods such as LoRA. Due to these options rise in computational cost, yet none is necessarily required for the present results. We therefore leave the pursuit of SOTA performance to future work and keep the focus on the substantive disclosure and corporate greenwashing. Besides, we acknowledge that DeepGreen has implemented specific optimizations to the text data tailored for the Chinese market. For international markets, it is equally essential to conduct targeted task optimizations that align with particular language types and structures. Fortunately, leveraging the inherent cross-language natural language processing capabilities of LLMS, such task adaptations will be facilitated. However, given that this paper centers on a Chinese perspective as its core narrative, we intend to integrate research pertaining to international markets into our future endeavors.}

\re{Equally important are our empirical findings. We document a robust positive association between corporate greenwashing and the subsequent likelihood of receiving an environmental penalty. The result survives a battery of sensitivity checks: alternative measurement for violations, one-year lag structure, PSM, and placebo test. Because the \textbf{Greenwashing} indicator is built from the theoretical attributes, the results that are greenwashing but unrelated to penalty in heterogeneity analysis point to the systemic neglect of regulatory agencies.}

\re{The prevailing enforcement mindset equates supervision with the punishment of visible violations, greenwashing that conceals those emissions is ultimately unraveled, yet the same penalty is simply imposed after the fact. It simultaneously bestows relative immunity on corporations that have stockpiled symbolic assets such as ISO certificates or favored by foreign investment. These assets act as a protective buffer so that in the short term their probability of sanction does not rise even though their underlying environmental performance needs greenwashing. Though regulators whose mandate stops at compliance therefore incur no direct cost from this asymmetry, the market does pay the price, because capital will continue to flow toward the corporations that green risk is most underpriced. Specifically, institutional investors commonly lack the authority to verify environmental performance independently, so they inevitably rely on part of the public signals. If those signals  themselves are the product of greenwashing, the market will invert the true ranking of these corporations. Once greenwashing corporations are absorbed into portfolio, the capital allocation begins to concentrate rather than spread risk. Consequently,  greenwashing exposure is more than a leftover regulatory gap, but also a systemic tail risk that will erupt once the gap between appearance and reality shuts, bringing chaos to the market.}

\re{In summary, this paper contributes in two practical dimensions. Firstly, it offers regulators a ready-to-use screening tool that distils greenwashing from A-share large-scale dataset in 1$\sim$2 hours. The pipeline couples unsupervised LLMs with a ESG performance and flags corporations whose talk exceed their walk. Our effectiveness verification work confirms the high credibility of the unsupervised text analysis results achieved by LLM, providing experimental data reference for regulators to use artificial intelligence tools. Secondly, the empirical evidence establishes greenwashing as a lead indicator of corporate violations. Investors can treat the indicator as a proxy for potential green risk and  an early warning. The heterogeneity results add a caution by showing that symbolic assets weaken the link between greenwashing and punishment and do not reduce the underlying propensity to greenwash. Recognising this shielding effect allows supervisors to move from ex-post penalty to ex-ante oversight, and to target preventive measures at the corporations whose environmental narratives have outrun their environmental performance before any material damage occurs.}

\section*{Declaration}

\subsection*{CRediT authorship contribution statement}
\noindent \textbf{Congluo Xu}: Conceptualization, Data curation, Formal analysis, Funding acquisition, Investigation, Methodology, Project administration, Resources, Software, Supervision,Validation, Writing - original draft, Writing - review \& editing. \textbf{Jiuyue Liu}: Formal analysis, Investigation, Methodology, Validation, Writing - review \& editing. \textbf{Ziyang Li}: Investigation, Project administration, Supervision, Writing - review \& editing. \textbf{Chengmengjia Lin}: Investigation, Software, Visualization, Writing - review \& editing.

\subsection*{Availability of Data and Material}
\noindent The dataset used in this study includes 9369 annual reports of 3123 Chinese A-share listed companies from 2021 to 2023, from publicly available website \textit{Juchao} (www.cninfo.com.cn). The data used in the empirical study comes from CNRDS and \textit{Huazheng}'s official website.

\section*{Acknowledgements}
\noindent We thank \textbf{Yu Miao} (School of Economics, Sichuan University) and \textbf{Yiling Xiao} (Business School, Sichuan University) for their contributions to the early research and the original manuscript.

\newpage
\bibliographystyle{apalike}  
\bibliography{references}

@article{mohanram2025does,
  title={Does financial information presentation format matter? Evidence from Chinese firms’ reporting of research and development expense},
  author={Mohanram, Partha and Sun, Wei and Xin, Baohua and Zhu, Jigao},
  journal={Review of Accounting Studies},
  volume={30},
  number={2},
  pages={1638--1682},
  year={2025},
  publisher={Springer}
}

@article{clapham2023policy,
  title={Policy making in the financial industry: A framework for regulatory impact analysis using textual analysis},
  author={Clapham, Benjamin and Bender, Micha and Lausen, Jens and Gomber, Peter},
  journal={Journal of Business Economics},
  volume={93},
  number={9},
  pages={1463--1514},
  year={2023},
  publisher={Springer}
}

@article{dyer2017evolution,
  title={The evolution of 10-K textual disclosure: Evidence from Latent Dirichlet Allocation},
  author={Dyer, Travis and Lang, Mark and Stice-Lawrence, Lorien},
  journal={Journal of Accounting and Economics},
  volume={64},
  number={2-3},
  pages={221--245},
  year={2017},
  publisher={Elsevier}
}

@article{harris1954distributional,
  title={Distributional structure},
  author={Harris, Zellig S},
  journal={Word},
  volume={10},
  number={2-3},
  pages={146--162},
  year={1954},
  publisher={Taylor \& Francis}
}

@article{lyon2015means,
  title={The means and end of greenwash},
  author={Lyon, Thomas P and Montgomery, A Wren},
  journal={Organization \& environment},
  volume={28},
  number={2},
  pages={223--249},
  year={2015},
  publisher={Sage Publications Sage CA: Los Angeles, CA}
}

@article{bernini2024measuring,
  title={Measuring greenwashing: A systematic methodological literature review},
  author={Bernini, Francesca and Giuliani, Marco and La Rosa, Fabio},
  journal={Business Ethics, the Environment \& Responsibility},
  volume={33},
  number={4},
  pages={649--667},
  year={2024},
  publisher={Wiley Online Library}
}

@article{vaswani2017attention,
  title={Attention is all you need},
  author={Vaswani, Ashish and Shazeer, Noam and Parmar, Niki and Uszkoreit, Jakob and Jones, Llion and Gomez, Aidan N and Kaiser, {\L}ukasz and Polosukhin, Illia},
  journal={Advances in neural information processing systems},
  volume={30},
  year={2017}
}

@article{achiam2023gpt,
  title={Gpt-4 technical report},
  author={Achiam, Josh and Adler, Steven and Agarwal, Sandhini and Ahmad, Lama and Akkaya, Ilge and Aleman, Florencia Leoni and Almeida, Diogo and Altenschmidt, Janko and Altman, Sam and Anadkat, Shyamal and others},
  journal={arXiv preprint arXiv:2303.08774},
  year={2023}
}

@article{zhang2025siren,
  title={Siren’s Song in the AI Ocean: A Survey on Hallucination in Large Language Models},
  author={Zhang, Yue and Li, Yafu and Cui, Leyang and Cai, Deng and Liu, Lemao and Fu, Tingchen and Huang, Xinting and Zhao, Enbo and Zhang, Yu and Chen, Yulong and others},
  journal={Computational Linguistics},
  pages={1--46},
  year={2025},
  publisher={MIT Press 255 Main Street, 9th Floor, Cambridge, Massachusetts 02142, USA~…}
}

@article{hua2024limitations,
  title={The limitations and ethical considerations of chatgpt},
  author={Hua, Shangying and Jin, Shuangci and Jiang, Shengyi},
  journal={Data intelligence},
  volume={6},
  number={1},
  pages={201--239},
  year={2024},
  publisher={MIT Press One Rogers Street, Cambridge, MA 02142-1209, USA journals-info~…}
}

@article{gorovaia2025identifying,
  title={Identifying greenwashing in corporate-social responsibility reports using natural-language processing},
  author={Gorovaia, Nina and Makrominas, Michalis},
  journal={European Financial Management},
  volume={31},
  number={1},
  pages={427--462},
  year={2025},
  publisher={Wiley Online Library}
}

@article{zou2025esgreveal,
  title={ESGReveal: An LLM-based approach for extracting structured data from ESG reports},
  author={Zou, Yi and Shi, Mengying and Chen, Zhongjie and Deng, Zhu and Lei, ZongXiong and Zeng, Zihan and Yang, Shiming and Tong, Hongxiang and Xiao, Lei and Zhou, Wenwen},
  journal={Journal of Cleaner Production},
  volume={489},
  pages={144572},
  year={2025},
  publisher={Elsevier}
}

@article{bronzini2024glitter,
  title={Glitter or gold? Deriving structured insights from sustainability reports via large language models},
  author={Bronzini, Marco and Nicolini, Carlo and Lepri, Bruno and Passerini, Andrea and Staiano, Jacopo},
  journal={EPJ Data Science},
  volume={13},
  number={1},
  pages={41},
  year={2024},
  publisher={Springer Berlin Heidelberg}
}

@article{ziolo2024literature,
  title={Literature review of greenwashing research: State of the art},
  author={Zio{\l}o, Magdalena and B{\k{a}}k, Iwona and Spoz, Anna},
  journal={Corporate Social Responsibility and Environmental Management},
  volume={31},
  number={6},
  pages={5343--5356},
  year={2024},
  publisher={Wiley Online Library}
}

@article{liu2023greenwashing,
  title={Why greenwashing occurs and what happens afterwards? A systematic literature review and future research agenda},
  author={Liu, Yupei and Li, Weian and Wang, Lixiang and Meng, Qiankun},
  journal={Environmental Science and Pollution Research},
  volume={30},
  number={56},
  pages={118102--118116},
  year={2023},
  publisher={Springer}
}

@article{cai2016corporate,
  title={Corporate environmental responsibility and firm risk},
  author={Cai, Li and Cui, Jinhua and Jo, Hoje},
  journal={Journal of Business Ethics},
  volume={139},
  number={3},
  pages={563--594},
  year={2016},
  publisher={Springer}
}

@article{boedijanto2024potentials,
  title={Potentials and challenges of artificial intelligence-supported greenwashing detection in the energy sector},
  author={Boedijanto, Felice Janice Olivia and Delina, Laurence L},
  journal={Energy Research \& Social Science},
  volume={115},
  pages={103638},
  year={2024},
  publisher={Elsevier}
}

@article{acheampong2024social,
  title={Do social and environmental disclosures impact information asymmetry?},
  author={Acheampong, Albert and Elshandidy, Tamer},
  journal={Economics Letters},
  volume={234},
  pages={111487},
  year={2024},
  publisher={Elsevier}
}

@article{lyon2011greenwash,
  title={Greenwash: Corporate environmental disclosure under threat of audit},
  author={Lyon, Thomas P and Maxwell, John W},
  journal={Journal of economics \& management strategy},
  volume={20},
  number={1},
  pages={3--41},
  year={2011},
  publisher={Wiley Online Library}
}

@article{walker2012harm,
  title={The harm of symbolic actions and green-washing: Corporate actions and communications on environmental performance and their financial implications},
  author={Walker, Kent and Wan, Fang},
  journal={Journal of business ethics},
  volume={109},
  number={2},
  pages={227--242},
  year={2012},
  publisher={Springer}
}

@article{szabo2021perceived,
  title={Perceived greenwashing: the effects of green marketing on environmental and product perceptions},
  author={Szabo, Szerena and Webster, Jane},
  journal={Journal of business ethics},
  volume={171},
  number={4},
  pages={719--739},
  year={2021},
  publisher={Springer}
}

@article{de2020concepts,
  title={Concepts and forms of greenwashing: A systematic review},
  author={de Freitas Netto, Sebasti{\~a}o Vieira and Sobral, Marcos Felipe Falc{\~a}o and Ribeiro, Ana Regina Bezerra and Soares, Gleibson Robert da Luz},
  journal={Environmental Sciences Europe},
  volume={32},
  number={1},
  pages={19},
  year={2020},
  publisher={Springer}
}

@article{kim2015greenwash,
  title={Greenwash vs. brownwash: Exaggeration and undue modesty in corporate sustainability disclosure},
  author={Kim, Eun-Hee and Lyon, Thomas P},
  journal={Organization science},
  volume={26},
  number={3},
  pages={705--723},
  year={2015},
  publisher={Informs}
}

@article{parguel2015can,
  title={Can evoking nature in advertising mislead consumers? The power of ‘executional greenwashing'},
  author={Parguel, B{\'e}atrice and Benoit-Moreau, Florence and Russell, Cristel Antonia},
  journal={International Journal of Advertising},
  volume={34},
  number={1},
  pages={107--134},
  year={2015},
  publisher={Taylor \& Francis}
}

@article{jones2019rethinking,
  title={Rethinking greenwashing: Corporate discourse, unethical practice, and the unmet potential of ethical consumerism},
  author={Jones, Ellis},
  journal={Sociological Perspectives},
  volume={62},
  number={5},
  pages={728--754},
  year={2019},
  publisher={Sage Publications Sage CA: Los Angeles, CA}
}

@article{seele2017greenwashing,
  title={Greenwashing revisited: In search of a typology and accusation-based definition incorporating legitimacy strategies},
  author={Seele, Peter and Gatti, Lucia},
  journal={Business strategy and the environment},
  volume={26},
  number={2},
  pages={239--252},
  year={2017},
  publisher={Wiley Online Library}
}

@article{li2023effects,
  title={Effects of greenwashing on financial performance: Moderation through local environmental regulation and media coverage},
  author={Li, Wei and Li, Weining and Sepp{\"a}nen, Veikko and Koivum{\"a}ki, Timo},
  journal={Business strategy and the environment},
  volume={32},
  number={1},
  pages={820--841},
  year={2023},
  publisher={Wiley Online Library}
}

@article{cao2022carbon,
  title={Carbon information disclosure quality, greenwashing behavior, and enterprise value},
  author={Cao, Qilin and Zhou, Yunhuan and Du, Hongyu and Ren, Mengxi and Zhen, Weili},
  journal={Frontiers in Psychology},
  volume={13},
  pages={892415},
  year={2022},
  publisher={Frontiers Media SA}
}

@article{xu2023unveiling,
  title={Unveiling the “Veil” of information disclosure: Sustainability reporting “greenwashing” and “shared value”},
  author={Xu, Wei and Li, Mingzhu and Xu, Sen},
  journal={PLoS One},
  volume={18},
  number={1},
  pages={e0279904},
  year={2023},
  publisher={Public Library of Science San Francisco, CA USA}
}

@article{ashforth1990double,
  title={The double-edge of organizational legitimation},
  author={Ashforth, Blake E and Gibbs, Barrie W},
  journal={Organization science},
  volume={1},
  number={2},
  pages={177--194},
  year={1990},
  publisher={INFORMS}
}

@article{delmas2011drivers,
  title={The drivers of greenwashing},
  author={Delmas, Magali A and Burbano, Vanessa Cuerel},
  journal={California management review},
  volume={54},
  number={1},
  pages={64--87},
  year={2011},
  publisher={Sage Publications Sage CA: Los Angeles, CA}
}

@article{yu2020greenwashing,
  title={Greenwashing in environmental, social and governance disclosures},
  author={Yu, Ellen Pei-yi and Van Luu, Bac and Chen, Catherine Huirong},
  journal={Research in international business and finance},
  volume={52},
  pages={101192},
  year={2020},
  publisher={Elsevier}
}

@article{christensen2022corporate,
  title={Why is corporate virtue in the eye of the beholder? The case of ESG ratings},
  author={Christensen, Dane M and Serafeim, George and Sikochi, Anywhere},
  journal={The Accounting Review},
  volume={97},
  number={1},
  pages={147--175},
  year={2022},
  publisher={American Accounting Association}
}

@article{bochkay2023textual,
  title={Textual analysis in accounting: What's next?},
  author={Bochkay, Khrystyna and Brown, Stephen V and Leone, Andrew J and Tucker, Jennifer Wu},
  journal={Contemporary accounting research},
  volume={40},
  number={2},
  pages={765--805},
  year={2023},
  publisher={Wiley Online Library}
}

@article{martin2016managers,
  title={Managers’ green investment disclosures and investors’ reaction},
  author={Martin, Patrick R and Moser, Donald V},
  journal={Journal of Accounting and Economics},
  volume={61},
  number={1},
  pages={239--254},
  year={2016},
  publisher={Elsevier}
}

@article{bingler2022cheap,
  title={Cheap talk and cherry-picking: What ClimateBert has to say on corporate climate risk disclosures},
  author={Bingler, Julia Anna and Kraus, Mathias and Leippold, Markus and Webersinke, Nicolas},
  journal={Finance Research Letters},
  volume={47},
  pages={102776},
  year={2022},
  publisher={Elsevier}
}

@article{marquis2016scrutiny,
  title={Scrutiny, norms, and selective disclosure: A global study of greenwashing},
  author={Marquis, Christopher and Toffel, Michael W and Zhou, Yanhua},
  journal={Organization science},
  volume={27},
  number={2},
  pages={483--504},
  year={2016},
  publisher={Informs}
}

@article{zhou2024threat,
  title={Threat or shield: Environmental administrative penalties and corporate greenwashing},
  author={Zhou, Kuo and Qu, Zhi and Liang, Jiayang and Tao, Yunqing and Zhu, Mengting},
  journal={Finance Research Letters},
  volume={61},
  pages={105031},
  year={2024},
  publisher={Elsevier}
}

@article{yue2023media,
  title={Media attention and corporate greenwashing behavior: Evidence from China},
  author={Yue, Jun and Li, Yilin},
  journal={Finance Research Letters},
  volume={55},
  pages={104016},
  year={2023},
  publisher={Elsevier}
}

@article{li2025annual,
  title={Annual report audit, ESG report assurance and audit quality: Evidence from the same accounting firm},
  author={Li, Wenwen and Li, Ting and Zhu, Hongjun},
  journal={China Journal of Accounting Research},
  volume={18},
  number={3},
  pages={100434},
  year={2025},
  publisher={Elsevier}
}

@article{cohen2015nonfinancial,
  title={Nonfinancial information preferences of professional investors},
  author={Cohen, Jeffrey R and Holder-Webb, Lori and Zamora, Valentina L},
  journal={Behavioral research in accounting},
  volume={27},
  number={2},
  pages={127--153},
  year={2015},
  publisher={American Accounting Association}
}

@article{grigoras2024importance,
  title={The importance of the financial audit in the identification of fraud and errors recorded by companies},
  author={Grigoras-Ichim, Claudia-Elena and Bordeianu, Otilia-Maria and Morosan-Danila, Lucia},
  journal={Management},
  volume={43},
  number={2},
  pages={225--240},
  year={2024}
}

@article{wang2025construction,
  title={Construction and analysis of corporate greenwashing index: a deep learning approach},
  author={Wang, Xiao and Gao, Xukuo and Sun, Meng},
  journal={EPJ Data Science},
  volume={14},
  number={1},
  pages={44},
  year={2025},
  publisher={Springer}
}

@article{wei2023does,
  title={Does the “Greenwashing” and “Brownwashing” of Corporate Environmental Information Affect the Analyst Forecast Accuracy?},
  author={Wei, Jing},
  journal={Sustainability},
  volume={15},
  number={14},
  pages={11461},
  year={2023},
  publisher={MDPI}
}

@article{mu2023greenwashing,
  title={Greenwashing in corporate social responsibility: A dual-faceted analysis of its impact on employee trust and identification},
  author={Mu, Honglei and Lee, Youngchan},
  journal={Sustainability},
  volume={15},
  number={22},
  pages={15693},
  year={2023},
  publisher={MDPI}
}

@article{lubloy2025quantifying,
  title={Quantifying firm-level greenwashing: A systematic literature review},
  author={Lubl{\'o}y, {\'A}gnes and Kereszt{\'u}ri, Judit Lilla and Berlinger, Edina},
  journal={Journal of Environmental Management},
  volume={373},
  pages={123399},
  year={2025},
  publisher={Elsevier}
}

@article{calamai2025corporate,
  title={Corporate Greenwashing Detection in Text--a Survey},
  author={Calamai, Tom and Balalau, Oana and Guenedal, Th{\'e}o Le and Suchanek, Fabian M},
  journal={arXiv preprint arXiv:2502.07541},
  year={2025}
}

@article{sneideriene2025uncovering,
  title={Uncovering Greenwashing: Investigating Impression Management Gap in Corporate Reporting},
  author={Sneideriene, Agne and Legenzova, Renata},
  journal={Sustainability},
  volume={17},
  number={18},
  pages={8342},
  year={2025},
  publisher={MDPI}
}

@article{li2024does,
  title={Does social responsibility reform curb corporate greenwashing: Evidence from a quasi-natural experiment in China},
  author={Li, Ziyang and Luo, Tao and Li, Jiangyi and Tian, Yihao},
  journal={International Review of Financial Analysis},
  volume={96},
  pages={103623},
  year={2024},
  publisher={Elsevier}
}

@article{berg2022aggregate,
  title={Aggregate confusion: The divergence of ESG ratings},
  author={Berg, Florian and K{\"o}lbel, Julian F and Rigobon, Roberto},
  journal={Review of Finance},
  volume={26},
  number={6},
  pages={1315--1344},
  year={2022},
  publisher={Oxford University Press}
}

@article{shi2024effect,
  title={The effect of executive green human capital on greenwashing},
  author={Shi, Daqian and Lu, Shan and Fang, Ziwei},
  journal={Research in International Business and Finance},
  volume={71},
  pages={102461},
  year={2024},
  publisher={Elsevier}
}

@article{mateo2022international,
  title={An international empirical study of greenwashing and voluntary carbon disclosure},
  author={Mateo-M{\'a}rquez, Antonio J and Gonz{\'a}lez-Gonz{\'a}lez, Jos{\'e} M and Zamora-Ram{\'\i}rez, Constancio},
  journal={Journal of Cleaner Production},
  volume={363},
  pages={132567},
  year={2022},
  publisher={Elsevier}
}

@article{bridgeman2012comparison,
  title={Comparison of human and machine scoring of essays: Differences by gender, ethnicity, and country},
  author={Bridgeman, Brent and Trapani, Catherine and Attali, Yigal},
  journal={Applied Measurement in Education},
  volume={25},
  number={1},
  pages={27--40},
  year={2012},
  publisher={Taylor \& Francis}
}

@inproceedings{amorim2018automated,
  title={Automated essay scoring in the presence of biased ratings},
  author={Amorim, Evelin and Can{\c{c}}ado, Marcia and Veloso, Adriano},
  booktitle={Proceedings of the 2018 Conference of the North American Chapter of the Association for Computational Linguistics: Human Language Technologies, Volume 1 (Long Papers)},
  pages={229--237},
  year={2018}
}

@article{taboada2011lexicon,
  title={Lexicon-based methods for sentiment analysis},
  author={Taboada, Maite and Brooke, Julian and Tofiloski, Milan and Voll, Kimberly and Stede, Manfred},
  journal={Computational linguistics},
  volume={37},
  number={2},
  pages={267--307},
  year={2011},
  publisher={MIT Press One Rogers Street, Cambridge, MA 02142-1209, USA journals-info~…}
}

@article{kroon2024advancing,
  title={Advancing automated content analysis for a new era of media effects research: The key role of transfer learning},
  author={Kroon, Anne and Welbers, Kasper and Trilling, Damian and van Atteveldt, Wouter},
  journal={Communication Methods and Measures},
  volume={18},
  number={2},
  pages={142--162},
  year={2024},
  publisher={Taylor \& Francis}
}

@article{cheng2025impact,
  title={The impact of climate policy uncertainty on corporate ESG greenwashing},
  author={Cheng, Zhonghua and Wu, Yixuan},
  journal={Journal of Environmental Management},
  volume={394},
  pages={127353},
  year={2025},
  publisher={Elsevier}
}

@article{rohlfs2025generalization,
  title={Generalization in neural networks: A broad survey},
  author={Rohlfs, Chris},
  journal={Neurocomputing},
  volume={611},
  pages={128701},
  year={2025},
  publisher={Elsevier}
}

@article{ong2025towards,
  title={Towards Robust ESG Analysis Against Greenwashing Risks: Aspect-Action Analysis with Cross-Category Generalization},
  author={Ong, Keane and Mao, Rui and Varshney, Deeksha and Cambria, Erik and Mengaldo, Gianmarco},
  journal={arXiv preprint arXiv:2502.15821},
  year={2025}
}

@article{persakis2025greenwashing,
  title={Greenwashing in marketing: a systematic literature review and bibliometric analysis},
  author={Persakis, Antonios and Nikolopoulos, Theodoros and Negkakis, Ioannis C and Pavlopoulos, Athanasios},
  journal={International Review on Public and Nonprofit Marketing},
  pages={1--36},
  year={2025},
  publisher={Springer}
}

@article{liu2024deepseek,
  title={Deepseek-v3 technical report},
  author={Liu, Aixin and Feng, Bei and Xue, Bing and Wang, Bingxuan and Wu, Bochao and Lu, Chengda and Zhao, Chenggang and Deng, Chengqi and Zhang, Chenyu and Ruan, Chong and others},
  journal={arXiv preprint arXiv:2412.19437},
  year={2024}
}

@article{team2025kimi,
  title={Kimi k2: Open agentic intelligence},
  author={Team, Kimi and Bai, Yifan and Bao, Yiping and Chen, Guanduo and Chen, Jiahao and Chen, Ningxin and Chen, Ruijue and Chen, Yanru and Chen, Yuankun and Chen, Yutian and others},
  journal={arXiv preprint arXiv:2507.20534},
  year={2025}
}

@article{madaan2023self,
  title={Self-refine: Iterative refinement with self-feedback},
  author={Madaan, Aman and Tandon, Niket and Gupta, Prakhar and Hallinan, Skyler and Gao, Luyu and Wiegreffe, Sarah and Alon, Uri and Dziri, Nouha and Prabhumoye, Shrimai and Yang, Yiming and others},
  journal={Advances in Neural Information Processing Systems},
  volume={36},
  pages={46534--46594},
  year={2023}
}

@article{pang2024iterative,
  title={Iterative reasoning preference optimization},
  author={Pang, Richard Yuanzhe and Yuan, Weizhe and He, He and Cho, Kyunghyun and Sukhbaatar, Sainbayar and Weston, Jason},
  journal={Advances in Neural Information Processing Systems},
  volume={37},
  pages={116617--116637},
  year={2024}
}

@article{lewis2020retrieval,
  title={Retrieval-augmented generation for knowledge-intensive nlp tasks},
  author={Lewis, Patrick and Perez, Ethan and Piktus, Aleksandra and Petroni, Fabio and Karpukhin, Vladimir and Goyal, Naman and K{\"u}ttler, Heinrich and Lewis, Mike and Yih, Wen-tau and Rockt{\"a}schel, Tim and others},
  journal={Advances in neural information processing systems},
  volume={33},
  pages={9459--9474},
  year={2020}
}

@inproceedings{salemi2025comparing,
  title={Comparing retrieval-augmentation and parameter-efficient fine-tuning for privacy-preserving personalization of large language models},
  author={Salemi, Alireza and Zamani, Hamed},
  booktitle={Proceedings of the 2025 International ACM SIGIR Conference on Innovative Concepts and Theories in Information Retrieval (ICTIR)},
  pages={286--296},
  year={2025}
}

@article{yuan2024exaggerating,
  title={Exaggerating, distracting, or window-dressing? An empirical study on firm greenwashing recognition},
  author={Yuan, Xueying and Xu, Jinhua and Shang, Lixia},
  journal={Finance Research Letters},
  volume={67},
  pages={105845},
  year={2024},
  publisher={Elsevier}
}

@article{zhang2025encouraging,
  title={Encouraging or inhibiting: Can analyst attention reduce corporate greenwashing behavior?},
  author={Zhang, Mengzhi and He, Wenjian},
  journal={Economic Analysis and Policy},
  volume={85},
  pages={943--962},
  year={2025},
  publisher={Elsevier}
}

@article{chen2024mere,
  title={Mere facade? Is greenwashing behaviour lower in low-carbon corporates?},
  author={Chen, Pengyu and Chu, Zhongzhu},
  journal={Business Strategy and the Environment},
  volume={33},
  number={5},
  pages={4162--4174},
  year={2024},
  publisher={Wiley Online Library}
}

@article{dubey2024llama,
  title={The llama 3 herd of models},
  author={Dubey, Abhimanyu and Jauhri, Abhinav and Pandey, Abhinav and Kadian, Abhishek and Al-Dahle, Ahmad and Letman, Aiesha and Mathur, Akhil and Schelten, Alan and Yang, Amy and Fan, Angela and others},
  journal={arXiv e-prints},
  pages={arXiv--2407},
  year={2024}
}

@article{yang2025qwen3,
  title={Qwen3 technical report},
  author={Yang, An and Li, Anfeng and Yang, Baosong and Zhang, Beichen and Hui, Binyuan and Zheng, Bo and Yu, Bowen and Gao, Chang and Huang, Chengen and Lv, Chenxu and others},
  journal={arXiv preprint arXiv:2505.09388},
  year={2025}
}

@article{hu2023green,
  title={The green fog: Environmental rating disagreement and corporate greenwashing},
  author={Hu, Xinwen and Hua, Renhai and Liu, Qingfu and Wang, Chuanjie},
  journal={Pacific-Basin Finance Journal},
  volume={78},
  pages={101952},
  year={2023},
  publisher={Elsevier}
}

@article{magness2006strategic,
  title={Strategic posture, financial performance and environmental disclosure: An empirical test of legitimacy theory},
  author={Magness, Vanessa},
  journal={Accounting, Auditing \& Accountability Journal},
  volume={19},
  number={4},
  pages={540--563},
  year={2006},
  publisher={Emerald Group Publishing Limited}
}

@article{brown1998public,
  title={The public disclosure of environmental performance information—a dual test of media agenda setting theory and legitimacy theory},
  author={Brown, Noel and Deegan, Craig},
  journal={Accounting and business research},
  volume={29},
  number={1},
  pages={21--41},
  year={1998},
  publisher={Taylor \& Francis}
}

@article{wu2020bad,
  title={Bad greenwashing, good greenwashing: Corporate social responsibility and information transparency},
  author={Wu, Yue and Zhang, Kaifu and Xie, Jinhong},
  journal={Management Science},
  volume={66},
  number={7},
  pages={3095--3112},
  year={2020},
  publisher={INFORMS}
}

@article{liu2025does,
  title={How does green investor entry affect corporate carbon performance? Evidence from China},
  author={Liu, Maotao and Fang, Xubing},
  journal={Renewable Energy},
  volume={244},
  pages={122748},
  year={2025},
  publisher={Elsevier}
}

@article{jiang2021green,
  title={Do green investors play a role? Empirical research on firms’ participation in green governance},
  author={Jiang, GS and Lu, JC and Li, WA},
  journal={Journal of Financial Research},
  volume={5},
  pages={117--134},
  year={2021}
}

@article{rodrigue2013environmental,
  title={Is environmental governance substantive or symbolic? An empirical investigation},
  author={Rodrigue, Michelle and Magnan, Michel and Cho, Charles H},
  journal={Journal of Business Ethics},
  volume={114},
  number={1},
  pages={107--129},
  year={2013},
  publisher={Springer}
}

@article{guo2021foreign,
  title={Foreign ownership and corporate social responsibility: Evidence from China},
  author={Guo, Mingyuan and Zheng, Chendi},
  journal={Sustainability},
  volume={13},
  number={2},
  pages={508},
  year={2021},
  publisher={MDPI}
}

@article{plumlee2015voluntary,
  title={Voluntary environmental disclosure quality and firm value: Further evidence},
  author={Plumlee, Marlene and Brown, Darrell and Hayes, Rachel M and Marshall, R Scott},
  journal={Journal of accounting and public policy},
  volume={34},
  number={4},
  pages={336--361},
  year={2015},
  publisher={Elsevier}
}

@article{esposito2025sustainability,
  title={Sustainability in energy companies under the lens of cultural pressures: when do we talk of greenwashing?},
  author={Esposito, Paolo and Doronzo, Emanuele and Riso, Vincenzo and Tufo, Massimiliano},
  journal={Corporate Social Responsibility and Environmental Management},
  volume={32},
  number={3},
  pages={3814--3831},
  year={2025},
  publisher={Wiley Online Library}
}

@article{tan2024faking,
  title={Faking for fortune: Emissions trading schemes and corporate greenwashing in China},
  author={Tan, Ruipeng and Cai, Qijun and Pan, Lulu},
  journal={Energy Economics},
  volume={130},
  pages={107319},
  year={2024},
  publisher={Elsevier}
}

@article{sun2024bank,
  title={Bank competition and firm greenwashing: Evidence from China},
  author={Sun, Yabin},
  journal={Finance Research Letters},
  volume={63},
  pages={105279},
  year={2024},
  publisher={Elsevier}
}

\appendix
\newpage
\section{Method Comparison}\label{comparison}
\begin{figure}[!htbp]
    \centering
    \caption{\centering A Comparison of Differences and Advantages}
    \label{fig2}
    \includegraphics[width=0.9\linewidth]{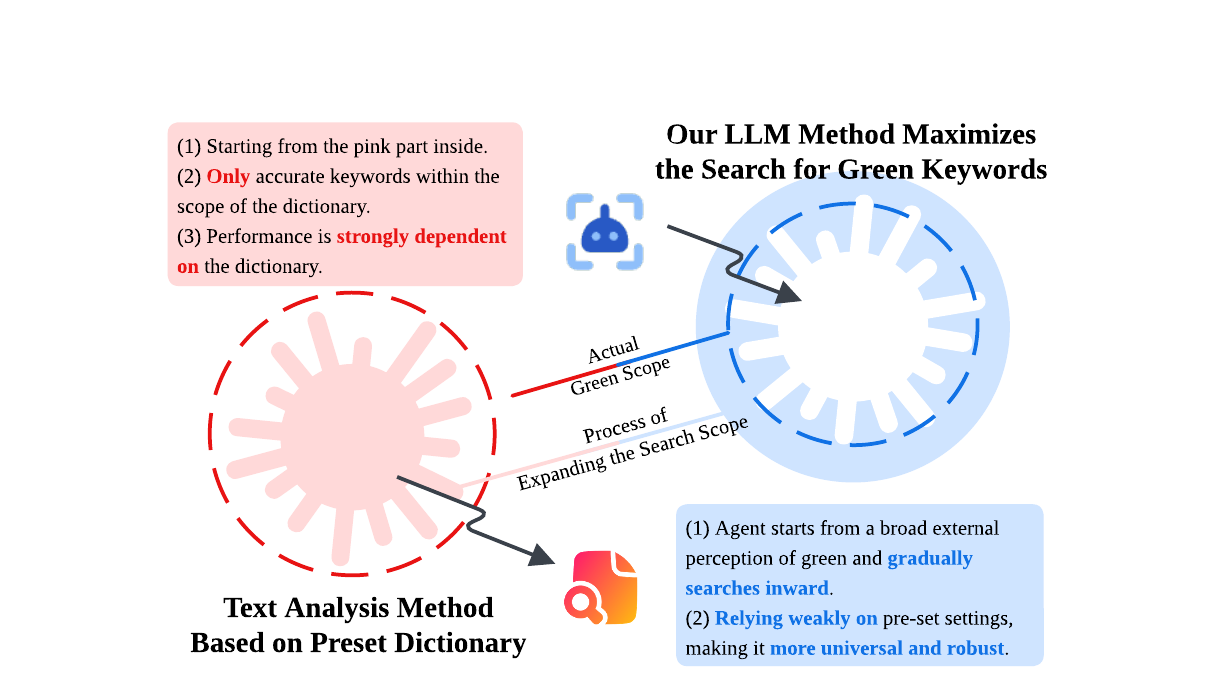}
\end{figure}

Our search method, as shown in \autoref{fig2}, outperforms traditional pre-set dictionary approaches. Given the rapid technological advancements and diverse environmental practices of companies, a single dictionary cannot effectively capture their unique green s. Incomplete dictionaries may skew word frequency data, amplifying issues in empirical study. Moreover, constructing a comprehensive dictionary is time-consuming, costly, and lacks timeliness. Our method, rooted in semantic segmentation, leverages public cognition to approximate professional understanding, ensuring that potential green s are not overlooked.

The severity of LLM judgment will be directly influenced by the user's intention. If the user wishes to make lenient judgments and explicitly express the intention of "lenient search" in the prompt, the range of keywords generated by the model will be relatively wide; On the contrary, if users want stricter judgments, the keyword range will be relatively narrow. This is an ideal state of human-computer interaction.

\newpage
\section{Task Templates}\label{templates}
\re{In this section, we will present the prompt template used for two tasks using LLM. As we are processing annual report text data in Chinese, all input will be in Chinese. However, for ease of understanding, we have translated it into English.}

\re{
\centering{\textbf{Layer A}}
\begin{itemize}
    \item \noindent \texttt{<Role Anchoring>:} You are an external ESG analyst who focuses most on the environmental aspects of corporate financial reporting ESG section.
    \item \noindent \texttt{<Persona Setting>:} {You are very familiar with the dimensions of corporate ESG, including but not limited to: green, environmental protection, environment, clean energy, energy conservation, carbon reduction, low-carbon, emission reduction, sewage treatment, air pollution, corporate environmental standards, national environmental regulations, etc.
    \item \noindent \texttt{<Task Description>:} You are an honest analyst and won't fabricate materials. You need to determine whether the given word is a keyword that related to corporate green disclosure.
    \item \noindent \texttt{<Answer Template>}:} Please strictly follow the JSON Schema format below and return a pure JSON string without any unnecessary explanations: \texttt{\{``judgment'':<int>, ``confidence'':<float>\}}, where $\mathrm{judgment=1}$ represents affirmative and 0 represents negative. The confidence level ranges from 0 to 1.
\end{itemize}}

\re{
\centering{\textbf{Layer B}}
\begin{itemize}
    \item \noindent \texttt{<Role Anchoring>:} You are an external ESG analyst who focuses most on the environmental aspects of corporate financial reporting ESG section.
    \item \noindent \texttt{<Persona Setting>:} {You are very familiar with the dimensions of corporate ESG, including but not limited to: green, environmental protection, environment, clean energy, energy conservation, carbon reduction, low-carbon, emission reduction, sewage treatment, air pollution, corporate environmental standards, national environmental regulations, etc.
    \item \noindent \texttt{<Task Description>:} You are an honest analyst and won't fabricate materials. You will be given in sequence: (1) a word; (2) a complete sentence containing that word (with the word marked by ``\texttt{\#\#}''). The type of word you receive mainly falls into the following four categories: (a) a word with book title marks, which is usually a certain document. You need to judge, in the specified context, whether this document is a document issued, responded to, or formulated by the corporation itself to solve its own and its subsidiaries’ environmental problems. For example, documents such as \textit{xx Corporation Environmental Project Assessment} and \textit{xx Corporation Pollution Prevention and Control Plan} indicate that this is the corporation’s voluntary disclosure of environmental information. However, documents such as \textit{Environmental Protection Law} and \textit{xx City Pollution Control Requirements} are documents issued by the state, government, and other non-corporation entities. Even if the corporation responds to this document, it is not the corporation’s voluntary disclosure of its own relevant environmental information. (b) a chemical substance or chemical term, which is usually a certain chemical term, such as ``manganese''``oxide''``condensation reaction'', etc. You need to judge, in the specified context, whether the corporation has dealt with the chemical substance through technical means or alleviated a certain type of environmental problem in the corporation’s production and operation process through this chemical means. (c) a word that will have a direct semantic association with environment, environmental protection, and green, such as ``Ministry of Environmental Protection''``green production''``carbon emission'', etc. You need to judge, in the specified context, whether the corporation has made a contribution to this word by issuing a specific document or adopting a specific technical method, or whether the corporation has alleviated a certain type of environmental problem in its production and operation process through this word. For example, ``The corporation responds to the carbon emission policy'' is just a slogan without actual action, while ``The corporation responds to the carbon emission policy and cleans the polluted gases'' includes actual action. (d) other words that should be associated with corporations and factories, such as ''big chimney''. You need to judge, in the specified context, whether the corporation has actually and specifically dealt with the polluted word or actually and specifically promoted the environmentally beneficial word.
    \item \noindent \texttt{<Answer Template>}:} Please strictly follow the JSON Schema format below and return a pure JSON string without any unnecessary explanations: \texttt{\{``judgment'':<int>, ``confidence'':<float>\}}, where $\mathrm{judgment=1}$ represents affirmative and 0 represents negative. The confidence level ranges from 0 to 1.
\end{itemize}}

\end{document}